\documentclass[]{fairmeta}

\usepackage{amssymb,amsthm,amsmath}
\usepackage{wrapfig}
\usepackage{mathtools}

 \usepackage{xspace}

\renewcommand{\phi}{\varphi}

\usepackage{algorithm}
\usepackage{algorithmic}
\usepackage{float}
\usepackage{listings} 
\usepackage{pmboxdraw} 
\usepackage{amssymb}   
\usepackage[T1]{fontenc}
\usepackage{lmodern}
\usepackage{graphicx}
\graphicspath{{assets/}}

\title{SemaClaw: A Step Towards General-Purpose Personal AI Agents through Harness Engineering}

\author{Ningyan Zhu, Huacan Wang\textsuperscript{$\dagger$}, Jie Zhou, Feiyu Chen, Shuo Zhang, Ge Chen, Chen Liu, Jiarou Wu, Wangyi Chen, Xiaofeng Mou, Yi Xu\textsuperscript{$\dagger$}}

\affiliation{Midea AIRC}

\abstract{
The rise of OpenClaw in early 2026 marks the moment when millions of users began deploying personal AI agents into their daily lives, delegating tasks ranging from travel planning to multi-step research. This scale of adoption signals that two parallel arcs of development have reached an inflection point. First is a paradigm shift in AI engineering, evolving from prompt and context engineering to harness engineering—designing the complete infrastructure necessary to transform unconstrained agents into controllable, auditable, and production-reliable systems. As model capabilities converge, this harness layer is becoming the primary site of architectural differentiation. Second is the evolution of human–agent interaction from discrete tasks toward a persistent, contextually aware collaborative relationship, which demands open, trustworthy and extensible harness infrastructure. We present \textbf{SemaClaw}, an open-source multi-agent application framework that addresses these shifts by taking a step towards general-purpose personal AI agents through harness engineering. Our primary contributions include a DAG-based two-phase hybrid agent team orchestration method, a PermissionBridge behavioral safety system, a three-tier context management architecture, and an agentic wiki skill for automated personal knowledge base construction.
}

\date{March 28, 2026 \\[0.5em] \textbf{GitHub:} \href{https://github.com/midea-ai/SemaClaw}{https://github.com/midea-ai/SemaClaw} \\[0.6em] \mbox{}}

\begin{document}

\begin{tikzpicture}[remember picture, overlay]
\node[anchor=south west, xshift=2cm, yshift=0.5cm] at (current page.south west) {%
    \quad
    \textsuperscript{\textdagger} Corresponding author.%
};

\end{tikzpicture}

\maketitle

\begin{tikzpicture}[remember picture, overlay]
\node[anchor=north east, xshift=-3.5cm, yshift=-12.5cm] at (current page.north east) {%
    \includegraphics[height=4.5em]{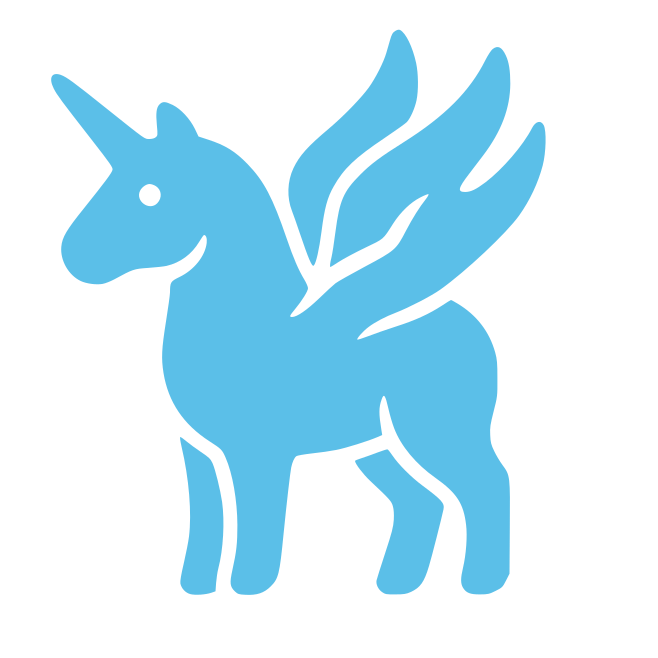}%
};
\end{tikzpicture}

\section{Introduction}

The rapid real-world adoption of open-source personal AI agents has surfaced a class of engineering challenges that model capability alone cannot resolve. OpenClaw~\cite{OpenClaw2026} is an instructive case: within weeks of release, hundreds of thousands of users were connecting it to messaging platforms, email, calendars, and file systems---delegating not just queries but consequential actions. This scale of deployment revealed that users need far more than a capable model. They need agents that can handle complex multi-step tasks reliably, operate within explicit safety boundaries, and accumulate useful knowledge across sessions. These are properties of the \textit{system around the model}---of harness design. Yet the open-source ecosystem for personal multi-agent applications has not converged on a framework that addresses them systematically. 

Three system-level challenges define this gap. \textit{First, real-world tasks often require dynamic yet structured orchestration.} Many user requests cannot be reduced to a fixed linear sequence of tool calls. Instead, they involve hierarchical decomposition, partial ordering among subtasks, intermediate dependency management, and localized recovery from failure. Existing approaches often occupy one of two extremes. On one side, declarative workflow systems~\cite{Lobster2026} provide explicit structure and observability, but are limited in their ability to adapt task structure at runtime. On the other side, unconstrained agentic reasoning offers flexibility, but frequently lacks execution traceability, reliable delegation, and robust failure isolation. In practice, developers attempting to build multi-agent behaviors on top of current agent frameworks often encounter pseudo-orchestration~\cite{Reddit2026}, in which a nominal ``orchestrator'' agent performs most reasoning internally instead of producing a verifiable and executable decomposition. This suggests the need for an orchestration mechanism that preserves runtime adaptability while maintaining explicit execution structure.

\textit{Second, as agents acquire the ability to perform consequential operations, behavioral safety must be enforced at the runtime level.} The safety problem in agent systems differs from conventional model safety. The central concern shifts from whether an agent generates harmful text to whether it executes actions that have not been explicitly authorized, such as modifying files, invoking external APIs, or running code. Existing open-source systems commonly treat permissions as application-level configuration or tool-level wrappers. Such approaches are insufficient once agents operate in open-ended environments where execution paths are dynamically determined. What is required is a runtime architecture in which authorization checkpoints are treated as first-class control primitives, rather than as optional safeguards attached to individual applications.

\textit{Third, sustained use of personal agents requires structured long-term memory, rather than simple accumulation of past dialogue.} Users interacting with an agent over weeks or months expect it to retain preferences, prior decisions, domain-specific conventions, and evolving background knowledge across sessions. However, many current memory mechanisms remain fundamentally log-oriented: they support retrieval of previous interactions, but do not adequately support the gradual consolidation of user- or task-specific knowledge into reusable conceptual structure. Prior work has explored important aspects of this space, including memory management inspired by operating-system paging in MemGPT~\cite{packer2023memgptllmsoperatingsystems} and reflective, hierarchical memory in generative agents~\cite{park2023generativeagentsinteractivesimulacra}. Nevertheless, open-source agent frameworks still lack a systematic design that combines short-term context control, retrieval-based external memory, and cross-session knowledge organization into a unified runtime abstraction. To bridge this gap, we argue that true persistence requires ``Knowledge Sedimentation''---externalizing task-derived insights into a durable, user-owned format.

These three challenges are not independent: they collectively define the engineering gap between a capable model and a production-reliable agent system---the gap that \textit{harness engineering} is concerned with closing.  

Production systems provide concrete evidence that harness design is a primary determinant of agent performance, independent of model capability. Claude Code~\cite{Anthropic2025claudecode} exemplifies mature harness engineering in the coding-agent domain: its explicit context lifecycle management, persistent state isolation, hook-based execution control, and incremental skill loading demonstrate how runtime architecture shapes agent reliability in practice. Empirically, LangChain's controlled experiments~\cite{LangChain2026} on Terminal Bench 2.0 show that holding the model constant while improving only harness configuration raises task completion from 52.8\% to 66.5\%---a 13.7 percentage point gain attributable entirely to harness design. Yet the corresponding design space remains underexplored in open-source frameworks for personal multi-agent applications, where orchestration, safety, and memory concerns must be addressed together rather than in isolation.


This gap motivates the primary mission of this work: to operationalize dynamic orchestration, runtime safety, and long-term memory within a robust, open-source harness architecture.

Toward this end, we present \textbf{SemaClaw}, 
a multi-agent application framework and an initial step toward general-purpose personal AI agents. SemaClaw is organized as a two-layer architecture: \texttt{sema-code-core}~\cite{SemaCodeCoreGithub}, an open-source event-driven agent runtime that manages context lifecycle, tool orchestration, and multi-tenant isolation, serves as the reusable foundation; \texttt{semaclaw} builds the application harness on top of it, providing channel integration, memory infrastructure extension, agent team coordination, and the plugin ecosystem.

\begin{figure}[H]           
    \centering              
    \includegraphics[width=\columnwidth]{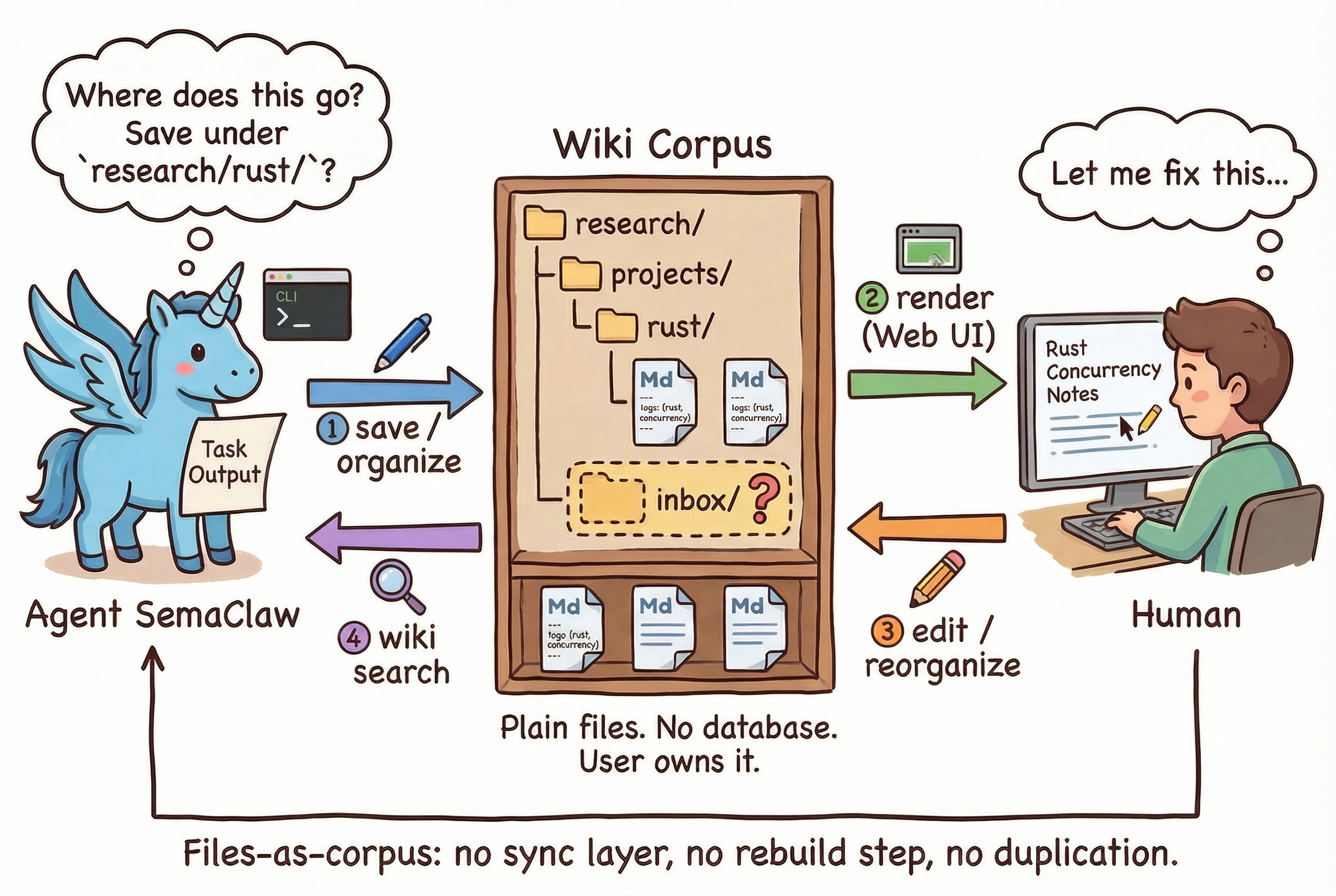}
    \caption{The wiki-based personal knowledge infrastructure as shared infrastructure between agent and user. The agent saves and organizes through a CLI; the user browses and edits through the Web UI; both retrieve from the same Markdown files, and edits made on either side are immediately visible to the other.}
    \label{fig:wiki}
\end{figure}

Built on this two-layer architecture, SemaClaw introduces a set of mechanisms that jointly address the three challenges outlined above as a coherent harness for personal AI agents.  For orchestration, we propose \textbf{DAG Teams}, a two-stage hybrid method in which an LLM first generates a task graph with explicit dependency edges, and a deterministic scheduler then executes the graph. This design combines the flexibility of dynamic task decomposition with the observability and fault locality of graph-structured execution. For behavioral safety, we introduce \textbf{PermissionBridge}, which embeds permission checkpoints into the runtime as execution primitives and requires explicit authorization before high-risk operations. For memory, we adopt a \textbf{three-layer context architecture} spanning compressed working memory, retrieval-based external memory, and a \texttt{SOUL.md}-anchored persona partition that maintains agent identity and behavioral alignment persistently across sessions. A dedicated wiki-based personal knowledge infrastructure complements this architecture, enabling structured accumulation and retrieval of user-specific knowledge as a personalized, durable knowledge base that feeds back into future agent sessions. As illustrated in Figure~\ref{fig:wiki}, a dedicated wiki-based personal knowledge infrastructure complements this architecture, enabling structured accumulation and retrieval of user-specific knowledge as a personalized, durable knowledge base that feeds back into future agent sessions. \textbf{Crucially, this relies on plain-file Markdown rather than opaque databases, ensuring the user retains absolute ownership over their compounding knowledge.}

In summary, this work makes the following contributions:

\begin{itemize}
    \item We present SemaClaw, an open-source multi-agent framework organized as a two-layer architecture that separates a reusable agent runtime from the application harness built on top of it.
    \item The framework incorporates a three-layer context management architecture that unifies working context, long-term memory retrieval, and per-agent persona partitioning.
    \item To close the trust gap between autonomous execution and user oversight, PermissionBridge is introduced as a native harness primitive that supports both explicit user authorization for high-risk tool actions and agent-initiated clarification requests.
    \item We adopt a four-layer plugin architecture—MCP tools, subagents, skills, and hooks—each anchored to a distinct engineering concern, as the principled extension surface of the SemaClaw harness.
    \item We propose DAG Teams, a two-stage hybrid orchestration framework that combines LLM-based dynamic task decomposition with deterministic DAG execution grounded in persistent agent personas.
    \item A four-mode scheduled task system—spanning pure notification, pure script, pure agent, and hybrid script-plus-agent execution—matches mode to task complexity to keep token consumption proportional to reasoning work.
    \item We develop a wiki-based personal knowledge infrastructure that externalizes task-derived knowledge into a user-owned, topic-organized corpus retrievable alongside agent memory—adding a dedicated knowledge layer through which what is learned can compound across future agent sessions.
\end{itemize}

\section{Technical Foundations}

\subsection{The ReAct Loop: From Single Inference to Agentic Execution}

The transition from language models as passive text generators to active agents capable of accomplishing multi-step tasks rests on a deceptively simple architectural insight: reasoning and acting can be interleaved within a single inference loop. The ReAct framework~\cite{yao2023react}, which formalized this pattern, defines an agent's execution as a repeating cycle of three operations:

\begin{lstlisting}
Thought → Action → Observation → Thought → ...
\end{lstlisting}

In each cycle, the model produces a \textit{Thought}—an explicit reasoning step that interprets the current state and determines what to do next. It then selects an \textit{Action}—a tool invocation that interacts with the external environment. The tool's return value becomes an \textit{Observation}, which is appended to the context and informs the next reasoning step. This cycle repeats until the agent determines that the task is complete or that it cannot proceed.

The significance of ReAct lies not in the individual operations but in their composition. By making reasoning steps explicit and interleaved with actions, the framework gives the agent a mechanism for self-correction: an observation that contradicts the agent's expectation can be incorporated into the next thought, redirecting execution without external intervention. This stands in contrast to the single-pass prompt engineering paradigm, where all reasoning must occur before any action is taken.

\subsubsection{Tool Use and the MCP Protocol}

The \textit{Action} step in the ReAct loop is only as powerful as the tools available to the agent. Early implementations described tools through natural language instructions within the prompt itself, relying on the model to parse tool signatures and format invocations correctly—a fragile approach that limited reliability in production settings. The introduction of structured function calling interfaces formalized this contract: tools are declared as typed schemas, invocations are machine-parseable, and return values are structured rather than free text.

The Model Context Protocol (MCP)~\cite{Anthropic2024mcp} further standardizes tool integration, allowing the agent's action space to be extended with external capabilities across frameworks and providers through a uniform interface.

\subsubsection{From Single Loop to Continuously Running Agent}

While the ReAct loop elegantly describes a single task execution, deploying agents in production introduces a category of problems the loop structure itself does not address.

\textbf{State persistence.} A ReAct loop is stateless between invocations: each new task begins with a fresh context. For agents intended to serve the same user across many sessions—accumulating knowledge, maintaining task continuity, and developing familiarity with the user's preferences—this statelessness is a fundamental limitation. Addressing it requires an explicit memory architecture that exists outside the context window and is selectively loaded at the start of each task.

\textbf{Context lifecycle management.} As a task progresses, the accumulating Thought–Action–Observation transcript consumes context window capacity. Without active management, the context either overflows (triggering hard truncation that silently discards earlier reasoning) or degrades in quality as the model's attention is diluted across an excessively long history. History compaction—the semantic compression of prior reasoning into a condensed summary—is the primary technique for managing this lifecycle.

\textbf{Multi-tenant isolation.} In deployments where a single agent runtime serves multiple users or groups concurrently, the state associated with each session must be strictly isolated. Leakage of context, memory, or tool state between tenants is both a correctness failure and a security failure. Achieving isolation in an asynchronous, concurrent execution environment requires runtime-level mechanisms—not merely application-level conventions.

These three problems—state persistence, context lifecycle management, and multi-tenant isolation—constitute the engineering gap between the ReAct loop as a research abstraction and a production-grade agent runtime.

\subsection{Context Management: Architecting the Agent's Cognitive Input}

Of the three engineering problems identified in Section 2.1.2, context lifecycle management is the one that exposes a deeper structural question: not how to build a faster buffer, but what information the agent should be reasoning over at each step of its execution. Answering that question requires a model of the agent's cognitive input—not just its message history.

Effective agentic behavior depends not only on the agent's reasoning capability but on the quality and structure of the information it reasons over. Context is not a passive accumulation of message history—it is the complete set of cognitive inputs available to the agent at each step of the ReAct loop~\cite{yao2023react}: the in-progress conversation, tool observations, retrieved long-term memory, and the stable constraints that define the agent's identity and task environment. The engineering problem is therefore not how to store more, but how to ensure that the right information, in the right form, reaches the agent at the right moment.

A larger context window does not resolve this problem. The \textit{lost-in-the-middle} phenomenon~\cite{Liu2024} demonstrates that model utilization degrades systematically when relevant information appears in the middle of a long context—regardless of window capacity. The binding constraint is not storage but \textit{cognitive density}: the ratio of task-relevant signal to noise in the context the agent actually reasons over.

Three structurally distinct sources of context contribute to that density. \textit{Working memory} is what resides in the context window at runtime—session-scoped, immediately accessible, and the direct input to every reasoning step. \textit{External memory} is persistent storage outside the window, accumulated across sessions and injected on demand through retrieval. \textit{Structured Context Injection} partitions the agent's stable identity from its current task environment, maintaining the distinction between what the agent \textit{is} and what it is \textit{currently doing}. These three sources differ in lifecycle, access latency, and governance requirements; conflating them into a single undifferentiated context produces systems that are fragile under task switching and prone to \textit{context rot}—the gradual degradation of reasoning quality as irrelevant or stale content accumulates in the window.

\subsubsection{The Context Window as a Managed Resource}

The context window is the agent's working memory: finite, high-value, and subject to rapid pollution under naive accumulation. Every \textit{Thought}–\textit{Action}–\textit{Observation} cycle appends to it—tool outputs, intermediate conclusions, exploratory reasoning that may have been superseded. Without active management, the window fills with low-density content that dilutes the model's attention and buries the constraints and decisions that actually govern the current task.

The failure mode is not simply overflow. Even before the window reaches capacity, reasoning quality degrades as the model's attention is distributed across an increasingly noisy history. Hard truncation—the fallback when no active management is in place—compounds the problem: it discards content by recency rather than relevance, silently removing early-session constraints that remain in force while retaining recent but low-value observations. The result is an agent that contradicts its own earlier commitments without any signal that something has gone wrong.

Active context management treats the window as a resource to be curated rather than a buffer to be filled. This means continuously distinguishing between content that must remain directly accessible—active constraints, pending decisions, the current task state—and content whose informational value has been absorbed and whose presence now contributes only noise. The three-source architecture described above is the structural expression of this discipline: each source is governed by the policy appropriate to its lifecycle, rather than all content competing for space in a single undifferentiated window.

\subsubsection{History Compaction and Context Lifecycle}

The working memory layer requires explicit lifecycle management because its content is both essential and perishable. Each reasoning step produces outputs that are immediately necessary—the next step cannot proceed without them—but that become progressively less valuable as the task advances. A tool observation that was critical three steps ago may be noise today; a constraint established at session start remains binding throughout.

Sliding-window truncation is the naive response to this pressure: keep the most recent N turns, discard the rest. The failure mode is predictable—early-session constraints and completed decision chains are discarded while recent but low-value content is retained. Truncation preserves recency, not relevance, and these are not the same.

\textit{History compaction} addresses this by treating context lifecycle as a semantic operation rather than a mechanical one. Rather than discarding prior history, compaction compresses it: the raw \textit{Thought}–\textit{Action}–\textit{Observation} transcript is replaced by a structured summary that preserves active constraints, open decisions, and completed-task state while discarding the exploratory process that produced them. The result is a high-density representation of what the agent needs to know to continue, not a verbatim record of how it got there. MemGPT~\cite{packer2023memgptllmsoperatingsystems} formalizes this intuition through an operating-system analogy: the context window functions as main memory, external storage as disk, and compaction as the paging mechanism that keeps the working set current without losing persistent state.

The tradeoffs are real. Compaction quality depends on the summarizing model's ability to identify what is load-bearing in the prior history—a judgment that requires understanding the task structure, not just the surface form of the conversation. A compaction that drops a constraint the user stated early in the session, or that misrepresents the outcome of a completed subtask, produces a corrupted working state that subsequent reasoning will build on without correction. The cost of compaction errors is therefore asymmetric: they are silent, they compound, and they cannot be recovered once the original history is gone. This asymmetry is what makes compaction a design problem rather than an implementation detail.

\subsubsection{External Memory and Retrieval Injection}

Working memory governs a single session; it cannot carry knowledge across the session boundary. Facts established in prior sessions—user preferences, project constraints, the outcomes of past tasks—are lost when the window is cleared, leaving the agent to rediscover them from scratch. For agents intended to serve the same user over weeks or months, this amnesia is a fundamental capability ceiling.

External memory addresses this by persisting information outside the context window and injecting it selectively when relevant. The governing principle is the same as in Retrieval-Augmented Generation (RAG)~\cite{DBLP:journals/corr/abs-2005-11401}: rather than loading all accumulated knowledge into the window at once, the system retrieves only what is relevant to the current task and injects it into context at the moment it is needed. The retrieval problem for agent memory differs from document retrieval in one important respect: the queries are not user-issued questions but inferred task states, and the relevant content is not factual knowledge but prior decisions, established preferences, and task history.

Retrieval strategy determines what the agent can recall and when. Sparse retrieval—keyword-based methods such as BM25—handles exact-match queries efficiently but degrades when the query and the stored content use different vocabulary for the same concept. Dense retrieval maps queries and documents into a shared semantic space, capturing conceptual similarity at the cost of exact-match precision. Hybrid approaches combine both signals, recovering the strengths of each at the expense of additional infrastructure. The choice among these strategies is not primarily a performance question but a coverage question: which classes of memory failure—missed recalls, false positives, cross-lingual gaps—are acceptable given the agent's task profile.

Retrieval timing introduces a second dimension of design. \textit{Pre-retrieval injection}—loading relevant memory into context before task execution begins, based on the task description or initial user message—ensures that established constraints and prior decisions are visible from the first reasoning step, not discovered mid-task when they may already have been violated. On-demand retrieval, triggered by the agent's own tool calls during execution, handles cases where relevance cannot be determined in advance. Most production systems use both: a pre-retrieval pass at task initiation followed by agent-initiated retrieval as the task unfolds.

The architecture described in Generative Agents~\cite{park2023generativeagentsinteractivesimulacra} illustrates how these components interact at scale: a \textit{memory stream} records all observations, a \textit{reflection} mechanism periodically synthesizes higher-level abstractions from raw entries, and a retrieval function scoring recency, importance, and relevance determines what is injected at each reasoning step. The lesson is not that agent memory systems should replicate this architecture, but that retrieval quality, injection timing, and memory organization are not separable concerns—they must be co-designed.

\subsubsection{Structured Context Injection}

The third source of context consists of inputs that are \textit{explicitly assembled and injected} by the harness at well-defined points in the agent's execution lifecycle, rather than emerging organically from interaction or retrieval.

Injection mechanisms fall into two broad categories. The first is \textit{file-based injection}, in which documents anchored to the agent's workspace or identity directory are loaded at session initialization. These files serve diverse semantic functions depending on the framework: \texttt{SOUL.md} encodes persistent agent persona in OpenClaw-style deployments; \texttt{AGENTS.md}~\cite{agentsmd} provides a general agent context interface describing available agents and their roles; \texttt{MEMORY.md} supplies a curated cross-session knowledge index; \texttt{RULES.md} specifies behavioral constraints and safety boundaries at the deployment level; \texttt{CLAUDE.md} encodes project-level conventions in coding-assistant contexts; and \texttt{Plan.md} externalizes task planning state for structured execution tracking. This design space is deliberately flexible: frameworks differ in which files they recognize, how they are named, and whether they are human-authored, agent-maintained, or both. What these files share is a common governance model: they are workspace-scoped and loaded as a coherent precondition rather than retrieved by query.

The second category is \textit{programmatic injection}, in which context is assembled dynamically through code rather than read from static documents. This includes system-level metadata such as current time, locale, and environment state; task-oriented structured preprocessing that assembles relevant context based on the inferred task type before the agent begins reasoning; and runtime-generated reminders that re-inject critical constraints after compaction boundaries. Unlike file-based injection, programmatic injection is typically invisible to the user and operates as an automatic harness function rather than a user-configured artifact.

Both categories share a design objective: to ensure that the right context is available to the agent at the right point in its execution, assembled and injected at deliberately chosen lifecycle boundaries. The failure mode of under-injection is an agent that must reconstruct relevant background from scratch at each reasoning step; the failure mode of over-injection is a context polluted with irrelevant or premature information that dilutes the model's attention. Structured injection, whether file-based or programmatic, is the harness mechanism through which this balance is actively governed.

\subsection{Plugin Ecosystems: Four Layers of Agent Capability Extension}

Having established how the agent's cognitive input is structured and governed, the next question is how that input space is populated and extended at runtime—what mechanisms determine which tools, knowledge, and behavioral constraints are available to the agent at any given moment.

A recurring tension in agent system design is the trade-off between generality and focus. An agent with access to every capability is one whose reasoning is perpetually burdened by irrelevant options; an agent with too narrow a capability set cannot serve the range of tasks users bring to it. The solution is not a fixed compromise but a \textit{layered extension architecture}: different mechanisms extend agent capabilities at different abstraction levels, each corresponding to a different concern.

We identify four such mechanisms, each operating at a different point in the agent's execution and cognitive structure.

\subsubsection{MCP Tools: Extending the Action Space}

The Model Context Protocol (MCP)~\cite{Anthropic2024mcp}—despite its name—is more precisely understood as a \textit{tool extension context protocol}: its purpose is not to manage the model's internal context window, but to standardize how external tools are declared, discovered, and invoked, allowing tool ecosystems to be composable across agent runtimes and permission policies to be applied consistently regardless of tool origin. 

MCP tools operate at the \textit{Action} step of the ReAct loop: each tool extends the set of operations the agent can perform in the external world—reading from a file system, querying a database, calling an API, sending a message. The boundary of what the agent can \textit{do} is precisely the union of its available tools.

The engineering challenges at this layer extend beyond interface design (tool schemas that are self-describing and unambiguous) and reliability (tools should fail gracefully and return structured errors the agent can reason about). A more structural tension arises from \textit{context overhead}: MCP tool definitions are injected into the agent's context at invocation time, and infrastructure-oriented tool providers routinely expose large tool surfaces—Playwright, for instance, introduces 32 distinct tool entries. As the number of integrated tool providers grows, the cumulative context cost can crowd out task-relevant information and degrade reasoning quality. This has motivated a directional shift: rather than registering every capability as a discrete MCP tool, an emerging pattern is to expose coarser-grained \textit{skills}—self-contained, goal-oriented procedures invocable via a single CLI entry point—that encapsulate multi-step tool sequences internally (Section 2.3.3). This approach trades fine-grained tool composability for context efficiency, and better matches the granularity at which agents actually reason about tasks.

\subsubsection{Subagents: Delegation Through Prompt-Defined Interfaces}

Where MCP tools extend what an agent can do in the world, subagents~\cite{Anthropic2024agents} extend what an agent can \textit{think}—by delegating subtasks to specialized agents whose reasoning is better suited to particular problem domains. Each subagent is not merely a sub-task executor but a first-class agent in its own right: it carries an independent persona, a dedicated system prompt, and its own workflow configuration, allowing its reasoning style, tone, and tool access to be tailored independently of the orchestrator. The interface between orchestrator and subagent is fundamentally a prompt: the orchestrator describes the task in natural language; the subagent interprets it according to its own persona and context.

This mechanism operates at the Prompt Engineering layer of the stack. The orchestrator need not know the internal implementation of the subagent—only its identity and the kinds of tasks it handles well. This semantic interface is both the strength and the risk of subagent delegation: it is flexible and human-readable, but it places the burden of correct task specification on the orchestrator's language generation rather than on a typed contract.

Context isolation is a second, underappreciated benefit of the subagent model. When a subtask is delegated to a subagent, that subtask executes within the subagent's own context window—separate from the orchestrator's. The orchestrator receives only the result, not the intermediate reasoning trace. This means that complex tasks decomposed across multiple subagents do not cause the orchestrator's context to grow proportionally with task complexity; the orchestrator maintains a high-level view while each subagent bears the local cognitive load of its assigned subtask. This property is what makes subagent delegation qualitatively different from simply inserting more reasoning steps into a single agent's loop: it is a mechanism for scaling to task complexity without degrading the orchestrator's reasoning quality.

The persistent-persona model discussed in Section 3.5 exploits both properties: each subagent's stable identity document serves as the semantic interface through which the orchestrator makes delegation decisions, while the subagent's isolated context ensures that the execution of complex, multi-step subtasks does not propagate cognitive load back to the orchestrator.

\subsubsection{Skills: Progressive Capability Loading}

Agent skills operate at the Context Engineering layer. Originally introduced by Anthropic as a first-class mechanism in Claude Code~\cite{Anthropic2025skills}, skills represent a principled approach to the composition of modular capabilities. A skill is a self-contained capability package—a combination of prompt instructions, context documents, and tool references—that is progressively injected into the agent's context on demand rather than loaded at startup.

This progressive loading is implemented through a two-level lazy injection mechanism. At the skill registry level, inactive skills are represented only by their \texttt{name} and \texttt{description} fields parsed from a YAML frontmatter header --- sufficient for the agent to reason about available capabilities without loading full skill content. At the intra-skill level, skills can structure their internal assets (prompt templates, reference documents, tool configurations) as named sections with explicit routing directives, allowing the agent to load only the asset subtree relevant to the current invocation rather than the entire skill package. 

The progressive loading approach has two complementary benefits. First, it keeps the agent's context lean: capabilities irrelevant to the current task do not consume context window space or introduce reasoning noise. Second, it allows capability extension without agent redeployment: new skills can be authored, distributed, and activated without modifying the agent runtime.

This pattern was introduced in Claude Code as a mechanism for domain-specific capability injection and has been adopted more broadly—including in ClawHub's skill distribution model—as a general-purpose approach to non-technical capability extension. The key design requirement for a skill system is that skill injection be \textit{deterministic and scoped}: the agent should be able to reason about which skills are relevant and what they enable, rather than encountering injected instructions as unexplained additions to its context.

\subsubsection{Hooks: Harness Clamp Points}

Hooks operate at the outermost layer—the Harness Engineering level—and are distinct from the preceding three mechanisms in a fundamental way: they are not extensions of what the agent can do or know, but rather \textit{insertion points through which the harness exerts behavioral control over the agent's execution}.

A hook is a callback registered against a named lifecycle event in the agent runtime: task start, tool invocation (before and after), permission request, message dispatch, task completion, or error. When the event fires, the hook executes—potentially inspecting the event payload, logging it, modifying it, blocking it, or triggering secondary actions in external systems.

The harness metaphor is most precise at this layer. Hooks are the clamps and guides that keep the agent's execution within intended boundaries without requiring the agent itself to be aware of the constraints. An audit log hook records every tool invocation for compliance purposes; a guardrail hook blocks file deletion actions in protected directories; an observability hook forwards state transitions to a monitoring dashboard. None of these require changes to the agent's reasoning; all are applied transparently at the runtime level.

\subsubsection{The Four Layers as a Design Framework}

The four mechanisms are not alternatives but complements, each addressing a distinct extension need:

\begin{table*}[h]
\centering\small\renewcommand{\arraystretch}{1.2}
\resizebox{\textwidth}{!}{ 
\begin{tabular}{p{0.14\linewidth} p{0.25\linewidth} p{0.23\linewidth} p{0.25\linewidth}}
\toprule
Mechanism & Primary Engineering Concern & Extension Target & Interface Type \\
\midrule
MCP Tools & Tool Use & Agent action space & Typed schema \\
Subagents & Prompt Engineering & Agent reasoning scope & Prompt \\
Skills & Context Engineering & Agent capability & Progressive context injection \\
Hooks & Harness Engineering & Agent execution control & Lifecycle callbacks \\
\bottomrule
\end{tabular}
}
\caption{Four complementary agent extension mechanisms, each addressing a distinct engineering concern.}\par
\label{tab:plugin-layers}
\end{table*}

A well-designed agent platform provides clear extension points at all four layers. Conflating them --- for instance, using prompt injection to achieve what a hook should handle, or using a monolithic tool to perform what a subagent should reason about --- produces systems that are harder to reason about, test, and maintain. One mechanism deliberately excluded from this taxonomy is the \textit{slash command} interface introduced in Claude Code~\cite{Anthropic2025claudecode}. Slash commands provide a concise, user-facing shorthand for invoking skills, triggering subagents, executing hook scripts, and accessing built-in runtime functions---and can compose these mechanisms in a single invocation. They are omitted from the four-layer model not because they lack utility, but because they are more precisely characterized as a \textit{human--agent interaction shortcut} rather than an independent extension primitive: a slash command does not introduce new harness capability, but provides an ergonomic entry point into capabilities the other four layers already provide. 

Taken together, these four mechanisms constitute the extension surface of the agent harness: they are not ad-hoc additions but the structured interfaces through which harness engineering concerns --- capability composition, reasoning delegation, execution control, and context management --- are addressed at the appropriate layer. The mapping established here serves as an organizing principle for the SemaClaw plugin architecture described in Section 3.4.

\subsection{Multi-Agent Orchestration: From Call Chains to Collaborative Teams}

The four extension mechanisms in Section 2.3 expand what a single agent can do and know—but they do not change the fundamental constraint that all of that capability is exercised within one context window. When tasks require concurrent expertise, parallel execution across independent workstreams, or a cognitive load that exceeds any single window's capacity, extending a single agent further is the wrong solution. The unit of composition must shift from capability to agent.

The question is not whether to distribute work across agents, but how to structure the coordination: who decides what each agent does, when, and in what order. Three principal orchestration models have emerged in current practice, each representing a different answer to that question—and a different set of trade-offs between flexibility, predictability, and operational complexity.

\subsubsection{Stateless Swarm}

The stateless swarm model reduces multi-agent coordination to a single primitive: \textit{handoff}. Each agent executes within its own context until it determines that a different agent is better suited to continue; at that point, it transfers control by returning the target agent as the result of a tool call. The receiving agent picks up with the shared context variables passed at the handoff boundary and proceeds independently. No central coordinator exists; the execution path emerges from the sequence of handoff decisions made by individual agents.

OpenAI's Swarm framework~\cite{OpenAI2024swarm}is the canonical reference implementation of this pattern. Its design is explicitly minimal—agents are defined by a system prompt and a set of tools, handoffs are just tool calls that return another agent, and the framework itself maintains no state between turns. The official positioning is instructive: Swarm is described as an educational framework for exploring multi-agent interfaces, not a production system. This self-imposed scope boundary reflects a genuine architectural constraint rather than false modesty. Without persistent state, an agent that has transferred control cannot observe what happens next; without a coordinator, there is no mechanism for detecting when the overall task has gone off track or for recovering from a failed handoff chain.

The stateless model's strengths are real: it is simple to reason about, easy to test in isolation, and composable—adding a new agent requires only defining its prompt and tools, not restructuring a workflow graph. These properties make it well-suited to short, well-defined task chains where each step's output is a sufficient input for the next. Its limitations become apparent under long-running coordination: context accumulated in one agent's reasoning is not automatically available to the agent that receives the handoff, and there is no shared state to fall back on when the handoff chain breaks.

\subsubsection{Explicit Graph Structure (DAG-Based Orchestration)}

Where the stateless swarm leaves coordination implicit in the sequence of handoff decisions, explicit graph-based orchestration makes it a first-class artifact. Frameworks such as LangGraph~\cite{LangChain2024langgraph} model the workflow as a directed graph: nodes represent computation steps—agent invocations, tool calls, or human checkpoints—and edges encode the control flow between them. Conditional edges implement branching logic by routing execution to different downstream nodes based on the current state. The graph is defined before execution begins; the runtime traverses it deterministically.

The practical consequence of this explicitness is a qualitatively different failure model. When a node fails, the failure is localized: the graph structure identifies exactly which step produced the error, what its inputs were, and which downstream nodes were blocked as a result. Testing becomes tractable because each node can be exercised independently with controlled inputs. Human-in-the-loop checkpoints—pausing execution at a specific graph node to await approval before proceeding—are straightforward to implement because the execution state is fully captured in the graph's shared state object and can be persisted across the pause. These properties make explicit graph orchestration well-suited to workflows where correctness and auditability matter more than flexibility: compliance processes, multi-step data pipelines, and any task where the structure of the work is stable enough to specify in advance.

The limitation is symmetric with the strength. A workflow graph can only express task structures that were anticipated at design time. When a task's structure depends on information that only becomes available at runtime—when the number of subtasks is unknown, when subtask dependencies emerge from intermediate results, or when the appropriate agent for a step cannot be determined without first executing a prior step—the graph must either be over-specified with conditional branches for every contingency, or the task must be restructured to fit the graph rather than the graph adapted to the task. This rigidity is not a failure of implementation but a consequence of the model's core commitment: predictability requires that the execution structure be known before execution begins.

\subsubsection{Orchestrator Dynamic Decision-Making}

The third model inverts the relationship between structure and execution. Rather than defining the workflow graph in advance, the system designates one agent—the \textit{orchestrator}—as a runtime decision-maker responsible for decomposing the task, selecting which agents to dispatch, and determining the order of operations based on the evolving state of the task. The dependency structure between subtasks is not declared; it is discovered implicitly through the orchestrator's reasoning as the task unfolds.

This approach has been explored in several multi-agent frameworks. AutoGen~\cite{Wu2024} provides a representative architecture: a \texttt{GroupChatManager} acts as the orchestrator, dynamically selecting the next agent to speak based on the conversation history and the current task state. The pattern generalizes beyond chat: any system where an LLM-based coordinator reads current context and decides what to do next—which agent to invoke, with what instructions, and in what sequence—is an instance of orchestrator dynamic decision-making.

The flexibility this model provides is genuine. Tasks whose structure cannot be fully specified in advance—open-ended research, multi-step problem solving where each step's output shapes the next, coordination across agents with heterogeneous and overlapping capabilities—are handled naturally, because the orchestrator can adapt its decomposition strategy as new information arrives. The objective dependency structure between subtasks exists whether or not it is explicitly declared; the orchestrator's reasoning discovers and respects it without requiring the designer to enumerate it upfront.

The costs are equally genuine. The orchestrator's reasoning path is not directly observable; when the overall task fails, diagnosing whether the failure originated in the orchestrator's decomposition, in a worker agent's execution, or in the handoff between them requires reconstructing the full reasoning trace after the fact. Errors in the orchestrator's decisions propagate to all downstream workers before they can be detected. And because the orchestrator's behavior is a function of its context—which grows and changes as the task proceeds—reproducing a failure for debugging purposes requires reproducing the exact context state at the moment the bad decision was made, which is rarely straightforward. As discussed in Section 2.3.2, subagent context isolation partially mitigates the propagation problem by containing each worker's execution within its own context window; but it does not address the orchestrator's own reasoning opacity.

\subsubsection{Open Question: Dynamic Reasoning vs. Explicit DAG}

The three models described above are not equally suited to all tasks, and the choice between them is not purely technical—it reflects a judgment about which failure modes are more acceptable in a given deployment context. Stateless swarms optimize for simplicity at the cost of coordination visibility. Explicit graph structures optimize for predictability at the cost of pre-specification burden. Dynamic orchestration optimizes for flexibility at the cost of debuggability.

The deeper tension is between two properties that production systems require simultaneously: the ability to handle tasks whose structure is not fully known in advance, and the ability to understand, test, and recover from failures when they occur. Current orchestration models treat these as a trade-off rather than a conjunction—gaining one means conceding the other.

One direction worth exploring is whether the two approaches can be composed rather than chosen between. An orchestrator that reasons dynamically about task structure could, rather than dispatching subtasks immediately, first produce an explicit representation of the dependency graph it has inferred—a draft execution plan that can be inspected, modified, and validated before execution begins. The deterministic graph executor would then run the plan, providing the failure isolation and reproducibility of explicit graph orchestration while the orchestrator's dynamic reasoning handles the task structures that cannot be pre-specified. Whether the overhead of this two-phase approach is justified by the improvement in debuggability, and whether LLM-generated execution plans are reliable enough to serve as the basis for deterministic execution, remain open questions.

A second dimension concerns the relationship between orchestration structure and agent identity. The stateless swarm and explicit graph models treat agents as interchangeable computational units, distinguished only by their current prompt and tool configuration. The persistent-persona model—in which each agent maintains a stable identity document that persists across sessions—introduces a different basis for orchestration: the orchestrator routes tasks based on semantic alignment between task requirements and agent identity, rather than on hard-coded dispatch logic or graph structure. Whether this semantic routing is more robust than structural routing under task distribution shift, and how it interacts with the debuggability challenges of dynamic orchestration, are questions that practical deployments are only beginning to surface. Section 3.5 describes how SemaClaw implements this approach and the design decisions it entails.

The above foundations frame the contributions of SemaClaw. Section 3 describes the architecture and design rationale behind each contribution in detail.

\section{SemaClaw: Design and Implementation}

\begin{quote}\itshape
SemaClaw is an open-source multi-agent application framework built on top of \texttt{sema-code-core}, a reusable agent runtime. Its design is inspired by OpenClaw—an existing open-source agent framework that established the foundational patterns for channel-based agent deployment and multi-group management. SemaClaw extends this foundation with a decoupled runtime architecture, a persistent-persona-based agent team model, a structured context management system, and several additional capabilities described in this section. Where OpenClaw provides the conceptual blueprint, SemaClaw's contribution lies in the architectural refinements and new mechanisms built on top of it. The sections that follow examine each of these contributions in detail, documenting both the design decisions and the engineering rationale behind them.
\end{quote}

\subsection{Layered Architecture: Decoupling Core from Claw}

\subsubsection{The Separation Principle}

\begin{figure}[t]
    \centering
    \includegraphics[width=\columnwidth]{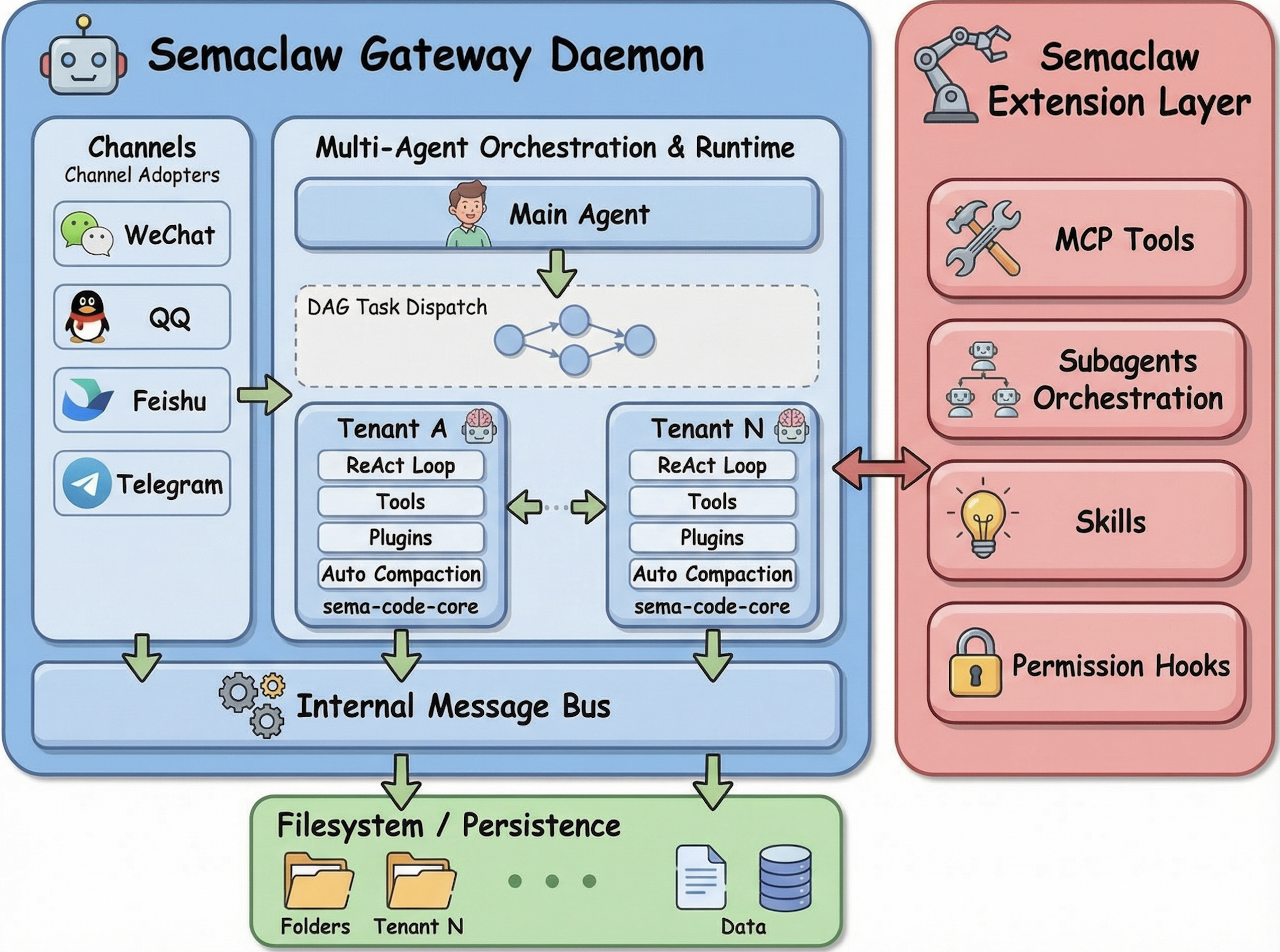}
    \caption{SemaClaw's two-layer architecture: \texttt{sema-code-core} provides the reusable agent runtime; \texttt{semaclaw} builds the application harness on top of it.}
    \label{fig:architecture}
\end{figure}

SemaClaw's architecture is organized as two independently released open-source projects with a strict dependency boundary between them, as illustrated in Figure~\ref{fig:architecture}.

\texttt{sema-code-core} provides the agent runtime: the execution loop, tool orchestration, context lifecycle management, and plugin interface. \texttt{semaclaw} provides the application layer built on top of it: the channel integration, message routing, agent pool management, memory infrastructure, plugin ecosystem, and agent team coordination.

The relationship is analogous to that between \texttt{pi-mono}\cite{pimonoGithub} and OpenClaw in the broader ecosystem: a runtime library that can be composed into application-specific frameworks, rather than a monolithic system in which runtime and application concerns are entangled.

\subsubsection{sema-code-core: An Event-Facade Runtime}

The design of \texttt{sema-code-core} follows an \textbf{event-facade pattern}: the runtime exposes a uniform event-driven interface through which all agent lifecycle transitions—session initialization, tool invocation, compaction triggers, context updates, session termination—are surfaced as typed events. Consumers of the runtime (including \texttt{semaclaw} itself) interact with the agent's execution through this facade rather than through direct access to internal state.

This architecture has two practical consequences. First, it enforces the boundary between runtime concerns and application concerns: the application layer responds to events and issues commands through the facade, but cannot reach into the runtime's internal execution state. Second, it makes the runtime composable: any application that understands the event protocol can be built on top of \texttt{sema-code-core} without modification to the runtime itself.

Alignment with Claude Code's design philosophy is intentional. Claude Code~\cite{Anthropic2025claudecode} represents a mature, production-tested approach to agentic execution—its conventions around context injection, history compaction, tool permission management, and session lifecycle have been validated at scale. By aligning \texttt{sema-code-core} with these conventions rather than inventing new ones, the runtime inherits a body of proven engineering decisions and provides a familiar foundation for developers already working in this ecosystem.

\subsubsection{Architecture Design Rationale}

The separation of runtime and application into distinct projects is not merely an organizational convenience—it reflects a substantive design claim about where the engineering concerns differ.

\textbf{Separation of concerns.} Agent runtime capabilities—executing ReAct loops, managing context windows, isolating tenant state—evolve on a different timescale and in response to different pressures than application orchestration logic. A new channel integration does not require changes to the execution loop; an improvement to history compaction does not require changes to message routing. Keeping these concerns in separate projects allows each to evolve at its own pace without forcing coupled releases.

\textbf{Reusability.} The runtime problems that \texttt{sema-code-core} solves—multi-tenant isolation, context lifecycle management, tool orchestration—are not specific to any particular application domain or channel. Any developer building an agent application faces these same foundational challenges. By extracting the runtime into an independent library, these solutions become available to the broader community as a reusable starting point rather than requiring each new project to solve them independently.

\textbf{Ecosystem composability.} The decoupled architecture creates a two-level extension surface: at the runtime level, developers can extend or replace individual runtime components while preserving the event-facade contract; at the application level, developers can build entirely new applications on top of the same runtime foundation. This mirrors the layered extension philosophy described in Section 2.3, applied to the architecture of the system itself.

\texttt{sema-code-core} is released as an independent open-source project. Developers seeking to build agent applications across any deployment context—messaging platforms, desktop assistants, API services, or otherwise—are encouraged to use it as a foundation, contributing to a shared runtime rather than maintaining isolated forks of the same core logic.

\subsection{Context Management in Practice}

Section 2.2 established working memory, external memory, and structured context injection as three sources of cognitive input; SemaClaw's layered context architecture instantiates these as in-context history with compaction, external persistent memory with hybrid retrieval, and persona-partitioned \textbf{soul} versus \textbf{workspace} directories as the structured injection layer. This section describes how that design maps onto running code. Each tier is governed independently, with its own storage substrate, lifecycle policy, and injection mechanism. 

\begin{figure}[t]           
    \centering              
    \includegraphics[width=\columnwidth]{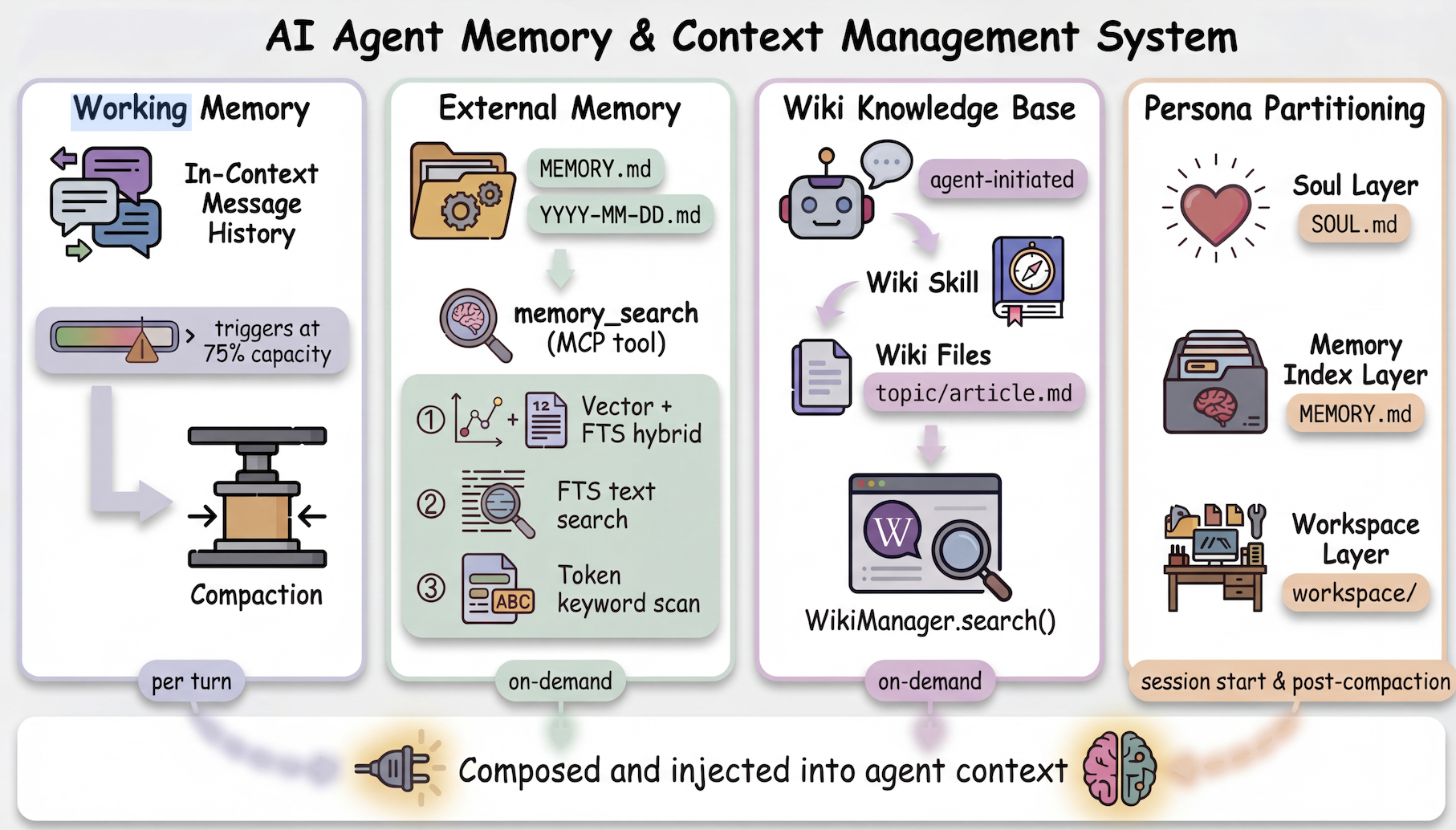}
    \caption{SemaClaw employs a three-layer context management architecture spanning working memory, external memory, and a SOUL.md-anchored persona partition, complemented by a wiki-based personal knowledge infrastructure that provides a personalized, durable knowledge base feeding back into retrieval. External Memory implements a three-level degradation strategy (Vector+FTS5 → FTS5 alone → Token keyword scan) for robust hybrid retrieval, ensuring availability even when vector search is unavailable or produces low-quality results.}
    \label{fig:memory_architecture}
\end{figure}

\subsubsection{Working Memory and Compaction}

Working memory in SemaClaw is owned entirely by \texttt{sema-code-core}. The application layer does not read or write the in-context message history directly; it interacts with it only through events. This boundary is deliberate: it keeps compaction logic—which requires intimate knowledge of the message format and token accounting—inside the runtime, while allowing the application layer to respond to compaction outcomes without depending on their internals.

Compaction is triggered automatically when the accumulated context reaches 75\% of the model's configured context length, with an 8,000-token overhead buffer applied to account for the next user message and injected reminders. When the threshold is crossed, \texttt{sema-code-core} emits a \texttt{compact:start} event before invoking the summarizing model, and a \texttt{compact:exec} event on completion—carrying the token counts before and after compression, the compression ratio, and the generated summary text. If the summarizing model call fails, a truncation fallback activates: the history is cut to 50\% of the context limit and a notice message is prepended, preserving at least a coherent recent window rather than leaving the agent with a corrupted state.

A non-obvious consequence of compaction is that the rules reminders injected at session start—the soul document, the cross-session memory index, and the project context—are no longer present in the compressed history. \texttt{sema-code-core} handles this by appending a new user-role message that embeds the output of \texttt{generateRulesReminders()} and, when the \texttt{TodoWrite} tool is enabled, \texttt{generateTodosReminders()}—ensuring that the agent's behavioral constraints and pending task state survive the compression boundary intact.

The application layer's role in this process is narrow but important. \texttt{AgentPool} subscribes to \texttt{compact:exec} and, on each firing, marks the current day's log file as dirty in \texttt{MemoryManager}. This signals that the file may have been written to since the last index pass and should be re-indexed before the next retrieval—ensuring that any conversation content recorded around the time of compaction is immediately searchable. Without this coupling, the external memory layer could serve stale results for the period between compaction and the next scheduled sync.

One current limitation is worth noting: the compaction prompt in \texttt{sema-code-core} was developed primarily against code-centric workflows. For general conversational agents—where the load-bearing content is preferences, relationship context, and open-ended commitments rather than code decisions and tool outputs—compression quality is less consistent, and this remains an area of active improvement.

\subsubsection{External Memory: Hybrid Retrieval and Injection}

Working memory is session-scoped and cleared at each compaction boundary; it cannot carry knowledge across sessions. External memory addresses this by persisting information outside the context window and making it selectively retrievable across the agent's operational lifetime.

Each agent's external memory occupies two primary file locations under its \texttt{agentDataDir}: \texttt{MEMORY.md}, a manually curated or agent-maintained index of persistent knowledge, and a rolling set of daily logs at \texttt{memory/YYYY-MM-DD.md}, written automatically by \texttt{DailyLogger} and retained for 50 days on a FIFO basis. Session-export JSON is a third file type anticipated by the architecture but not yet indexed by the current pipeline; \texttt{MEMORY.md} and the dated logs form the main on-disk corpus, both labeled \texttt{source: 'memory'} in the retrieval database—the distinction between them is semantic rather than structural: \texttt{MEMORY.md} accumulates curated, durable knowledge, while the dated logs capture the automatic conversation trace. The \texttt{memory\_search} tool exposes a \texttt{source} parameter (\texttt{memory}, \texttt{session}, or \texttt{all}) that filters by this label, allowing the agent to scope a query to curated-plus-log material, to rows tagged \texttt{session} when such corpora are present, or to the full index.

Retrieval is handled by \texttt{hybridSearch} in \texttt{MemoryManager}, which applies a three-level degradation strategy. When an embedding provider is configured, it combines FTS5 BM25 keyword scoring with vector similarity search against a \texttt{sqlite-vec} index. In the merged result set, both vector-only and FTS-only documents receive a base factor of 0.7; documents that appear in both paths receive \texttt{vec\_score × 0.7 + fts\_score × 0.3}. This symmetry is deliberate: FTS-only matches—common in cross-lingual queries via token expansion—would be systematically outranked if penalized to the lower 0.3 factor. If the vector search is unavailable or returns no results above the quality threshold, the system falls back to FTS alone; if FTS also fails, a token-level keyword scan serves as the final fallback. The indexing and query paths apply tokenization asymmetrically: documents are indexed without stopword filtering to preserve all retrievable terms, while queries are filtered to reduce noise—an engineering trade-off that favors recall at index time and precision at query time.

Memory is made available to the agent through an on-demand retrieval path: \texttt{memory\_search} is exposed as an MCP tool that the agent can invoke at any point during execution, with full control over query, result count, and source scope. This design places retrieval decisions within the agent's own reasoning loop—the agent determines when memory is relevant and queries accordingly, rather than receiving a fixed injection whose relevance the system must predict in advance. The daily log records only the original user prompt, not any retrieval results, so that the conversation trace reflects what the user said rather than what the system added.

\subsubsection{Persona Partitioning: Soul and Workspace}

External memory answers the question of what the agent knows across sessions; persona partitioning answers the complementary question of who the agent is—and what context the current task requires of it. The two tiers are stored separately, injected under distinct headings, and governed by different lifecycle policies precisely because conflating accumulated knowledge with stable identity creates fragile agents whose behavior shifts unpredictably as memory grows.

Persona partitioning is implemented as two structurally distinct injections assembled at each model invocation. The soul layer—resolved from the agent's fixed identity directory—carries the agent's stable persona (\texttt{SOUL.md}): its role, behavioral style, and persistent constraints. The workspace layer—resolved from the agent's current working directory—carries the task environment context: project conventions, domain knowledge, and workspace-specific instructions. A third component, the cross-session memory index (\texttt{MEMORY.md}), sits between them and provides the accumulated knowledge layer. The three components are injected under distinct headings, giving the model an unambiguous signal about the provenance of each piece of context: who the agent is, what it remembers, and what the current task requires.

The soul is fixed for the lifetime of an agent instance; the workspace is dynamic and can be switched at runtime without restarting the session or clearing the conversation history. This makes workspace switching a first-class operational primitive—an agent can move between project contexts while maintaining continuity of identity and memory. Each agent's memory and persona state is scoped to its own namespace, ensuring that concurrent agents share no retrieved content or injected context regardless of how many are running in the same process.

\subsection{Human-in-the-Loop: The PermissionBridge}

Agentic execution introduces a class of actions whose consequences are significant enough to warrant explicit human authorization before they proceed. A production harness must accommodate this requirement without treating human involvement as an exceptional interruption to the agent's loop—it must be a first-class primitive that the runtime handles gracefully, preserving session continuity across arbitrary approval delays.

\begin{wrapfigure}{r}{0.5\columnwidth} 
    \includegraphics[width=0.48\columnwidth]{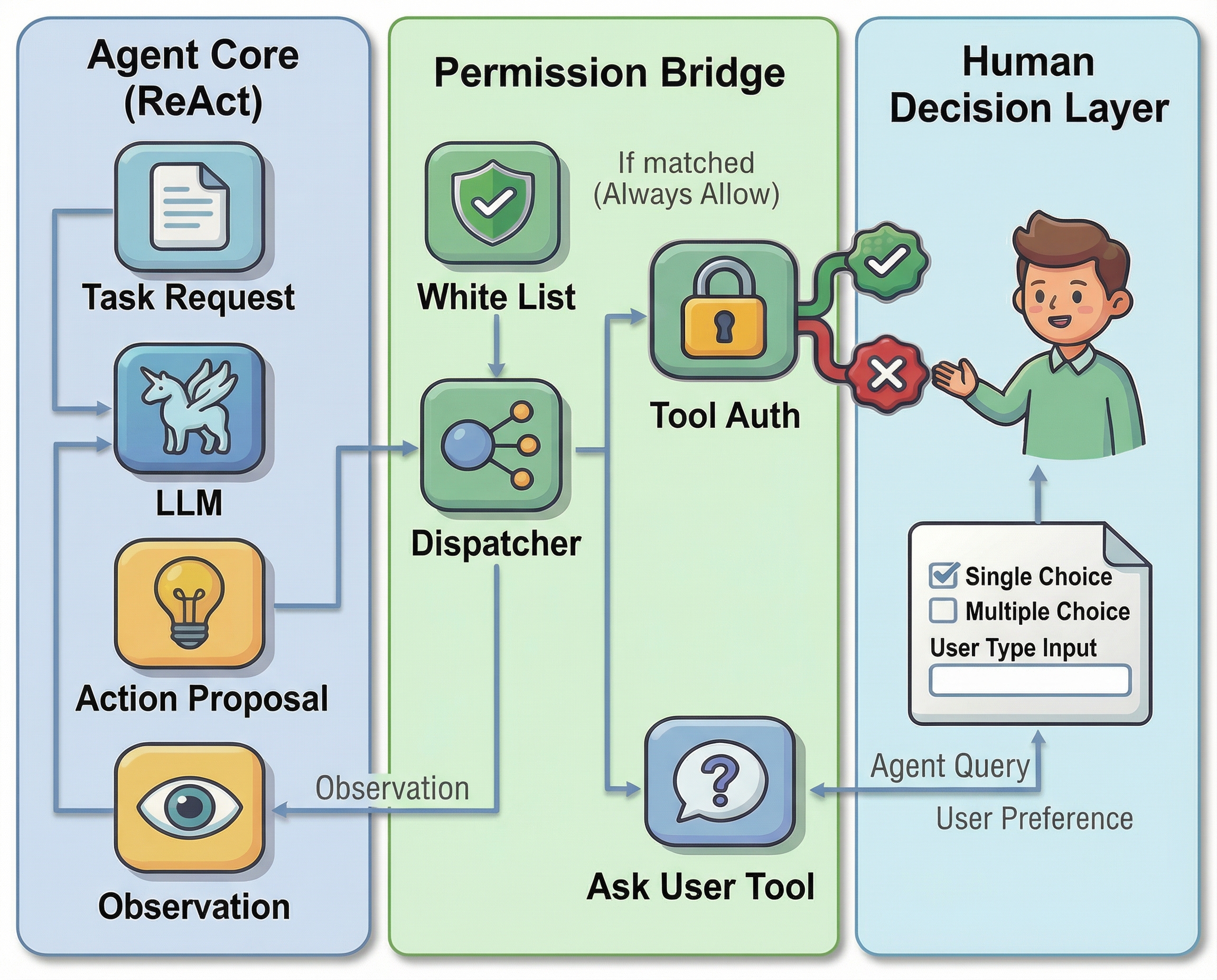}
    \caption{PermissionBridge coordination flow: both tool permission requests and agent-initiated user questions are routed through a single bridge instance, pausing execution until a response is received.}
    \label{fig:permission_bridge}
\end{wrapfigure}

SemaClaw addresses this through \texttt{PermissionBridge}, a globally scoped coordination layer that mediates between the agent runtime and the human approval interface, as illustrated in Figure~\ref{fig:permission_bridge}.

\subsubsection{Two Interaction Modes}

Human-in-the-loop interaction in SemaClaw takes two forms, corresponding to two distinct points in the agent's execution.

\textbf{Tool permission requests} arise when the agent attempts to invoke a tool that requires explicit user consent. The runtime pauses execution at the tool boundary, serializes the pending invocation—tool name, proposed arguments, and contextual rationale—and routes an approval request to the user. The user may approve, deny, or modify the request; on response, execution either resumes with the approved invocation or the agent receives a denial message it can reason about.

\textbf{User questions} arise when the agent determines that it lacks information necessary to proceed and requires clarification rather than authorization. The agent emits a structured question, execution pauses at the same boundary, and the user's response is returned to the agent as if it were a tool result—integrated directly into the next reasoning step.

Both modes follow the same coordination protocol, differ only in how the payload is presented to the user, and are routed through the same bridge instance.

\subsubsection{Permission Tiering}

Not all tool invocations carry the same risk profile, and requiring explicit approval for every tool call would make the agent unusable in practice. SemaClaw applies a two-tier permission policy.

Internal tools—memory retrieval, workspace management, inter-agent dispatch, and other infrastructure operations bundled with the runtime—are pre-authorized and invoke without user interaction. These tools operate entirely within the agent's own scope and cannot affect systems outside it; their risk profile does not warrant per-invocation approval.

External tools—user-installed MCP servers, file system operations, outbound API calls, and any tool that reaches beyond the agent's internal environment—require explicit per-invocation consent by default. The agent presents the proposed action and its arguments; the user approves or denies through whichever approval interface is available (channel inline buttons or Web UI). This boundary reflects the principle of least privilege applied at the tool layer: the agent holds no ambient permission to act on external systems; each such action must be explicitly sanctioned.

\subsubsection{The Bridge Design}

\texttt{PermissionBridge} operates as a single instance shared across all concurrent agent sessions. Concurrent approval requests are multiplexed by a unique request identifier assigned at the point of suspension; each response is routed back to the correct waiting execution context by matching this identifier. This design allows multiple agents to have simultaneous pending approvals without any coordination between them—the bridge handles fan-in and fan-out transparently.

The approval interface is channel-native: requests surface as interactive messages in the same Telegram conversation the user is already in, with inline action buttons for approve and deny. A parallel Web UI path is available for users who prefer a dashboard view. Both paths converge on the same bridge, so approval through either interface resumes the same waiting execution. The agent's session remains live throughout the approval window; the timeout mechanism operates in "last active" mode—any user interaction, including engagement with the approval prompt, extends the session lifetime, preventing approval latency from causing spurious session termination.

\subsubsection{Trust as a Design Primitive}

The PermissionBridge embodies a specific position on the relationship between human oversight and agent autonomy: the two are not in tension. A permission system that is bolted on after the fact—one that interrupts agent execution and requires special handling—encourages system designers to minimize its use. A permission system that is native to the runtime, with clear tiering and non-blocking behavior, makes human oversight the path of least resistance rather than an obstacle.

This design stance aligns with the broader harness engineering philosophy: constraints imposed structurally are more reliable than constraints imposed through convention. Requiring external tool invocations to pass through an approval gate is not a limitation on agent capability—it is the mechanism through which users extend the agent's trusted action space incrementally, on the basis of demonstrated judgment rather than assumed trustworthiness.

\subsection{Plugin Ecosystem in Practice}

SemaClaw's plugin ecosystem instantiates the four-layer model described in Section 2.3 (Table~\ref{tab:plugin-layers}) across each extension mechanism.

At the \textbf{MCP tool layer}, SemaClaw ships a suite of built-in servers that cover the agent's core operational needs: memory retrieval and indexing, workspace context management, inter-agent task dispatch, scheduled task management, and outbound message delivery. These internal servers are pre-authorized and require no user approval at invocation time. User-installed external MCP servers extend this base, subject to the permission policy described in Section 3.3.

At the \textbf{subagent layer}, the dispatch mechanism (described in Section 3.5) allows any agent designated as an orchestrator to delegate tasks to other named agents in the team. The subagent interface is prompt-defined: the orchestrator describes the task in natural language; routing is determined by persona alignment rather than explicit addressing.

At the \textbf{hooks layer}, lifecycle callbacks are available at key execution boundaries—task start, tool invocation, session compaction, and task completion—allowing external systems to observe or intervene in agent behavior without modifying the runtime.

At the \textbf{skills layer}, SemaClaw provides a skill management interface accessible through both the command-line interface and the Web UI. Users can browse available skills, install or remove them, and toggle individual skills on or off with immediate effect—no agent restart required. The Web UI skills page offers a visual catalog with per-skill descriptions and activation state; the CLI exposes the same operations for programmatic and headless environments. This dual-access model reflects the broader product goal of making agent capability management accessible regardless of the user's technical context.

\subsection{Agent Teams: Dynamic Orchestration on Persistent Personas}

\begin{figure}[H]           
    \centering              
    \includegraphics[width=\columnwidth]{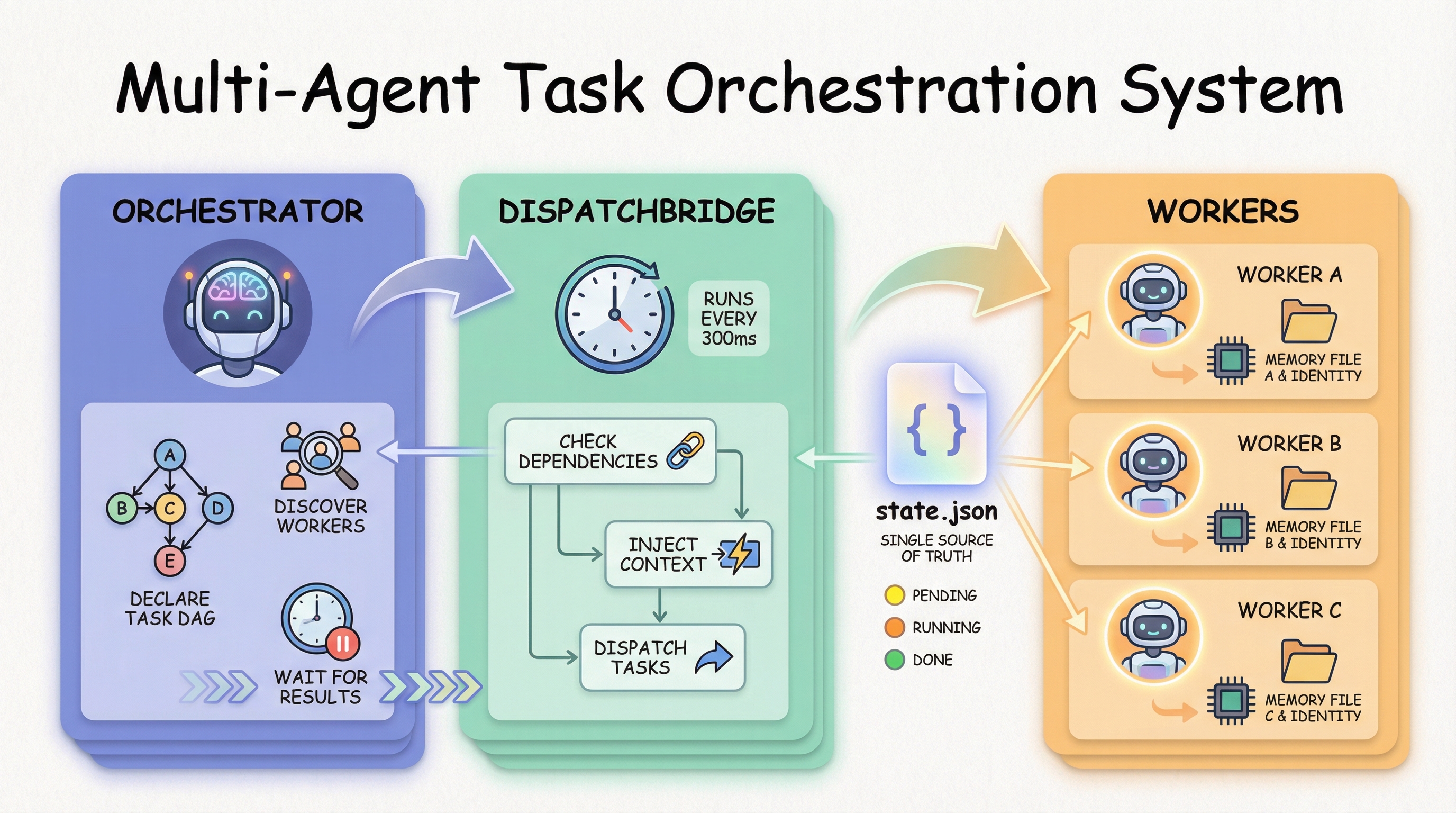}
    \caption{ SemaClaw implements a dynamic orchestration model where an orchestrator agent decomposes user tasks into subtasks with explicit dependencies (DAG), which the DispatchBridge runtime then dispatches to persistent worker agents for parallel execution. The architecture separates dynamic planning (LLM-driven task declaration by the orchestrator) from deterministic execution (scheduler-driven dispatch by DispatchBridge), ensuring both flexibility in task decomposition and debuggability in multi-agent workflows.}
    \label{fig:agent_team}
\end{figure}

Section 2.4.4 positioned persistent-persona routing—where an orchestrator selects workers based on semantic alignment with agent identity rather than hard-coded dispatch logic—as a distinct basis for orchestration, and noted that how it interacts with debuggability and robustness under distribution shift are questions practical deployments are only beginning to surface. This section describes how SemaClaw instantiates that approach in concrete engineering terms: how agent identities are established and maintained, how multi-agent tasks are declared and validated before execution begins, and how the runtime scheduler drives those tasks to completion.

\subsubsection{Persistent Personas as the Basis for Team Composition}

An agent team in SemaClaw is not assembled at runtime—it is a standing collection of persistent identities registered in the system. Each agent is identified by a \textit{folder} string that serves as its namespace across the entire stack: the \texttt{agentDataDir} holding its \texttt{SOUL.md}, \texttt{MEMORY.md}, and daily conversation logs is rooted at this folder, as is its default working directory. This identity is filesystem-level and session-independent; it persists across restarts and accumulates over the agent's operational lifetime.

When an agent is first registered, \texttt{ensureAgentDirs()} creates the full directory tree and writes an initial \texttt{SOUL.md} via \texttt{defaultSoulMd()}—seeding the agent's name, folder identifier, and default workspace path into its persona document. The file is written only once; all subsequent modifications are the user's or the agent's own. This initial document is the starting point for the soul resolution chain described in Section 3.2.3: the agent's stable identity is a file the system creates once and the operator shapes over time.

The orchestrator's task-routing decision rests on this identity layer, but the mechanism is precise. \texttt{list\_agents} exposes each registered worker's name, folder id, and channel to the orchestrator; the orchestrator's LLM reasoning determines which agent name to specify for a given subtask—a judgment that draws on those roster entries, the parent goal, and any persona or memory context the orchestrator already sees. The binding step, however, is deterministic: \texttt{resolveAgent()} performs a case-insensitive exact-string match against registered names and folder ids. There is no vector similarity at the binding layer. The semantic judgment happens entirely in the orchestrator's natural language reasoning; the resolution is a lookup. This separation—LLM decides who, string match confirms the binding—is what makes the routing both flexible and auditable.

Because each agent's memory accumulates under its folder, long-running agents develop specialization depth that stateless swarm participants cannot: a worker agent that has handled dozens of code-review tasks carries that history in its retrieval index, shaping the quality of its responses in ways that a freshly initialized agent cannot replicate.

\subsubsection{Task Declaration and the DAG Contract}

SemaClaw's dispatch path instantiates the composition Section 2.4.4 described: the orchestrator's LLM first emits an explicit dependency graph as structured data, and a deterministic executor in the runtime then runs that graph—rather than interleaving ad hoc dispatch decisions with each worker step.

The orchestrator declares a multi-agent task by calling \texttt{create\_parent} with a \texttt{goal} string and a \texttt{tasks} array. Each task entry carries an \texttt{agentName}, a \texttt{prompt}, and a \texttt{dependsOn} list of prior task labels—together forming an explicit directed acyclic graph of subtask dependencies. This declaration step happens before any worker is invoked: the orchestrator's LLM reasoning produces the full task structure as a single MCP call, not incrementally as execution proceeds. \texttt{detectCycle()} validates the declared graph at submission time; a cyclic dependency is rejected before it can cause a deadlock at runtime.

Two additional constraints govern the declaration. Only one parent task group per admin agent may be active at a time; additional \texttt{create\_parent} calls enter a \texttt{queued} state and are promoted sequentially as the active group completes. This serialization eliminates inter-group state contention without requiring the orchestrator to track which groups are in flight. When a parent is activated, the admin agent's current working directory is captured as \texttt{sharedWorkspace} and propagated to all workers in the group, giving the team a common file context when one is configured.

The result is that the orchestrator's dynamic reasoning is bounded to a single, well-defined moment: the construction of the task declaration. Everything after that point is governed by a deterministic scheduler with a known, inspectable state.

\subsubsection{Execution: The DispatchBridge as Deterministic Scheduler}

\texttt{DispatchBridge} runs in the main process and is responsible for driving the declared DAG to completion. Its core loop, \texttt{processPending()}, fires every 300 milliseconds and carries two responsibilities in a single pass. First, it scans all \texttt{processing} tasks for timeout expiry: any task whose \texttt{timeoutAt} has elapsed is marked \texttt{timeout} and removed from the active worker map. Second, it scans \texttt{registered} tasks whose dependencies have all reached a terminal state—\texttt{done}, \texttt{error}, or \texttt{timeout} all qualify—and dispatches them via \texttt{startTask()}. Treating failed upstream tasks as terminal ensures that downstream work is not permanently blocked by a single failure; the bridge advances the entire DAG rather than stalling at the first error.

\texttt{startTask()} constructs an augmented prompt for each worker that embeds the overall \texttt{<parent\_goal>}, the results of all \texttt{<prerequisites>} the current task depends on, and the status of \texttt{<other\_tasks>} in the same group. The worker therefore has full task context at the moment it begins execution, without needing to query the orchestrator mid-task. This is the practical expression of the subagent context isolation property discussed in Section 2.3.2: each worker's execution is self-contained, and the orchestrator's context does not grow with the cumulative reasoning of its workers.

The communication channel between the dispatch MCP server—which runs as a lightweight stdio subprocess—and \texttt{DispatchBridge} in the main process is a JSON state file, coordinated by a lock file for concurrent access. When a worker agent completes and its \texttt{sema-code-core} instance transitions to idle, \texttt{AgentPool} detects the state change and forwards the final reply to \texttt{DispatchBridge.notifyReply()}; on the error path, \texttt{notifyError()} writes the same terminal transition. Both paths call \texttt{processNextPending()} to immediately unblock any downstream tasks that were waiting on this worker, then conditionally restore the worker's original working directory—but only if no new task has been assigned to the same worker in the interim. A worker dispatched to consecutive tasks retains the shared workspace across them without an intermediate revert.

On startup, \texttt{DispatchBridge} marks any parents left in \texttt{active} or \texttt{queued} state from a previous run as \texttt{done}, with their in-flight tasks recorded as interrupted. A separate heartbeat, firing every two minutes, resets the admin agent's activity timer while it is blocked waiting on a long-running \texttt{dispatch\_task} poll—preventing the 30-minute no-activity timeout from triggering during legitimate multi-agent coordination.

The orchestrator waits on each subtask by calling \texttt{dispatch\_task}, which polls the state file at 500-millisecond intervals until the task reaches a terminal state. The polling deadline adjusts dynamically once a task begins executing, switching from the declaration-time timeout to the task's actual \texttt{timeoutAt}—avoiding the penalty that would otherwise fall on tasks that spent time waiting for upstream dependencies before they could start.

\subsection{A Four-Mode Scheduled Task System}

Scheduled tasks represent a distinct class of agent workload: they are triggered by time rather than by user input, and they range in complexity from a simple reminder that requires no model inference at all to a fully autonomous multi-step agent execution. A system that routes every scheduled task through a full agentic pipeline wastes both compute and token budget on work that does not require it; a system that only supports lightweight automation cannot address the use cases that benefit from agentic reasoning.

SemaClaw addresses this by defining four execution modes, each matched to a distinct complexity profile.

\textbf{Pure notification} delivers a time-triggered message to the user with no model invocation. The task is entirely deterministic: a scheduled time, a pre-authored message body, and a target channel. This mode is appropriate for reminders, recurring announcements, and any case where the content is known in advance and requires no adaptation at delivery time. It consumes no tokens and completes with minimal latency.

\textbf{Pure script} executes a deterministic code routine on schedule. The script runs outside the agent runtime—no context window, no model call, no tool permission overhead—and produces a structured output or side effect directly. This mode is appropriate for data collection, file processing, API polling, and other tasks whose logic is fully specifiable without LLM reasoning. Its primary advantages are stability (the behavior is exactly what the code specifies, with no stochastic variation) and cost efficiency (zero token consumption regardless of execution frequency or complexity).

\textbf{Pure agent} invokes a full agentic execution at a scheduled time. The agent receives a pre-authored prompt, reasons over its available tools and memory, and produces a response or takes actions autonomously. This mode is appropriate for tasks that require judgment, synthesis, or adaptation—summarizing the week's activity, drafting a status report, or initiating a research task—where the content cannot be determined in advance and benefits from the agent's full reasoning capability.

\textbf{Hybrid script-plus-agent} combines the two: a script runs first and produces structured data or pre-processed context, which is then passed to the agent as the input for its reasoning step. This mode addresses a common pattern in which the expensive part of a task is not reasoning but data gathering—fetching metrics, aggregating logs, querying APIs—and the agent's contribution is interpretation and synthesis rather than raw retrieval. By offloading the deterministic portions to a script, the hybrid mode keeps token consumption proportional to the reasoning work, not the total work.

The four modes reflect a deliberate design principle: \textbf{execution mode should be matched to task complexity}. An agent is not the right tool for every scheduled job, and treating it as such inflates cost, introduces unnecessary latency, and reduces system reliability. The taxonomy makes the trade-off explicit at configuration time, giving operators the vocabulary to deploy the right execution primitive for each use case.

\subsection{Wiki-Based Personal Knowledge Infrastructure: A User-Owned Knowledge Layer}

The preceding sections describe how SemaClaw executes tasks, manages context, and coordinates agents across time. None of these mechanisms, however, fully preserve what is \textit{learned} through interaction. The three-layer context architecture in Section~3.2 governs what the agent reads at each step, but the conclusions, distilled findings, and structured understanding produced during a task are still vulnerable to being absorbed into compaction summaries and eventually lost as logs roll off---well-suited to recall, ill-suited to knowledge sedimentation. Our key design insight is to externalize such learned knowledge into a user-owned corpus rather than leave it implicit inside session history. SemaClaw therefore adds a \textit{wiki-based personal knowledge infrastructure}: a dedicated knowledge layer that grows by deliberate curation rather than by accumulation, is organized by topic rather than by time, and remains directly legible and editable to the human rather than only to the agent. The wiki design is not incidental: a directory of plain Markdown files is simultaneously human-readable, version-controllable, and directly accessible to the agent---properties that a database or proprietary index cannot provide without introducing an intermediation layer that severs the user's direct ownership of their own knowledge.

\subsubsection{Mechanism: Storage, Construction, and Retrieval}

\textbf{Storage.} The knowledge layer is implemented as a directory tree of Markdown files on the user's local filesystem---each entry a Markdown body preceded by a YAML frontmatter header carrying tags and source metadata. There is no database and no proprietary index: the corpus is exactly what a file browser shows. This choice is deliberate. A user-owned knowledge layer should remain inspectable, movable, syncable, and editable without dependence on the running agent. The folder hierarchy \textit{is} the topic taxonomy, and the user can rename directories, restructure the tree, or move entries at any time without system involvement. The corpus is therefore a folder of files the user can edit, version-control, or migrate at will---with or without SemaClaw running.

\textbf{Construction.} The agent reaches this layer only through a small CLI operation set---inspect the tree, create a category, save an entry, organize an existing file---each returning structured output the agent can reason over. Two workflows are supported. \textit{Save} handles content the agent produces during a task: it inspects the tree, decides whether the content fits an existing category, creates a new one if not, or stages the entry in \texttt{inbox/} when classification is genuinely uncertain, then writes the entry with topic-relevant tags. \textit{Organize} handles user-supplied files: the agent copies them into the chosen category and edits only the frontmatter---adding tags and provenance---without rewriting the body. The constraint is structural rather than stylistic: the user's content is authoritative, and the agent's role is to classify and label, not to edit.

Category judgment is where the agent acts as a curator rather than a recorder. Recognizing that new content extends an established theme rather than warranting a new branch, or deferring to \texttt{inbox/} when the right call is ``I'm not sure yet''---these are reasoning operations, not I/O. They draw on the agent's reading of both the content and the user's accumulated taxonomy, and they shape the corpus into something a future retrieval pass and a human reader can both navigate.

\textbf{Retrieval.} The knowledge layer maintains its own search interface---content queries and tag filters over the corpus alone---distinct from the \texttt{memory\_search} tool of Section~3.2.2. The separation is intentional: external memory indexes what was \textit{said} (conversation transcripts, automatic logs), while the knowledge layer indexes what was \textit{learned} (content deliberately preserved as standalone knowledge). A \texttt{memory\_search} hit returns a slice of dialogue; a knowledge-layer search hit returns a self-contained, structured entry. Conflating them would erase the distinction that gives the layer its value. As a parallel retrieval source, it accumulates a body of curated, user-specific knowledge whose value compounds as the user's domain deepens.

\subsubsection{The Human Editorial Loop: Web UI and User-Side Organization}

A web interface renders the directory tree as a navigable knowledge base and each Markdown entry as a formatted document. The interface is not a layer built on top of the corpus---it is a view of the same files the agent reads and writes, with no intermediate state and no sync step. Through the UI, the user browses entries by category, edits content and frontmatter directly, and reorganizes the directory structure---renaming folders, moving entries, restructuring the taxonomy---without going through the agent. Any change is immediately visible to the next agent retrieval pass, because the files \textit{are} the corpus and there is no separate index to rebuild.

This files-as-corpus model closes the bidirectional loop between agent and user. Knowledge sedimented through agent task execution does not stay locked inside the agent---it surfaces as a corpus the user can read, correct, and extend. The agent's task work strengthens the user's own understanding: research the agent conducts becomes notes the user can study; decisions the agent reasoned through become a record the user can revisit. Every user refinement then flows back into the next retrieval pass, where the agent reads those edits as authoritative content. Over time the corpus becomes an artifact of joint authorship, growing in value to both parties through the same act of use.

\subsubsection{From Vibe Working to Vibe Learning}

\textbf{Vibe coding}, the original framing, compresses the distance between intent and code: the user expresses what they want, the agent produces it. \textbf{Vibe working} extends the same model beyond code to general task execution---drafting, research, analysis, coordination---making the agent a collaborator across the breadth of knowledge work. This knowledge layer introduces a third phase: \textbf{vibe learning}. Where vibe working helps users get things done, vibe learning ensures that getting things done leaves a residue---a structured, retrievable, growing record of what was done and what was understood through doing it.

Each completed task becomes an opportunity to consolidate the knowledge it required; each entry written is an investment in both the agent's future grounding and the user's own intellectual capital. This is the broader aspiration behind treating knowledge infrastructure as a first-class component of SemaClaw: not merely to build a better search surface, but to realize a system in which \textbf{intelligence compounds}. The knowledge accumulated through daily interactions---research conducted, decisions made, lessons extracted from failures---is organized, preserved, and made available to the agent and the user alike. In this framing, the agent's role is not to replace human cognition but to serve it: to extend the human capacity for recall, synthesis, and structured reflection beyond what any individual could sustain unaided. Section~4.4 returns to this design from a different angle, treating this knowledge layer as an instance of a stateful harness plugin whose effects persist beyond the session that produced them.

\section{Open Questions and Future Directions}

The sections above describe a system that works—but not one we consider finished. SemaClaw represents a particular set of answers to a set of engineering problems that are themselves still evolving. This section surfaces five questions that emerged during its design and deployment, where our current position is a working bet rather than a settled conclusion. We present each question with the tension we are navigating, the design choice we have made so far, and the directions we think are worth exploring further. We invite the community to stress-test these bets, propose alternatives, and share what they have learned in practice.

\subsection{Orchestration Architecture — Virtual Agents or Persistent Personas?}

\subsubsection{The Tension}

The multi-agent orchestration literature has produced two broadly different answers to a foundational question: \textit{what is an agent, exactly, in the context of a team?}

The first answer treats agents as \textbf{virtual participants}—ephemeral, lightweight, and role-scoped. An orchestrator assembles a team from scratch at task time, spinning up specialized instances defined by a role string or system prompt, using them for the duration of the task, and discarding them when the work is done. OpenAI's Swarm\cite{OpenAI2024swarm} is the clearest expression of this philosophy: agents are functions with metadata, with no identity that persists beyond the execution that created them. The appeal is composability and simplicity: a new role requires only a new prompt, and there is no state to manage between runs.

The second answer treats agents as \textbf{persistent collaborators}—enduring identities with accumulated memory, established working styles, and stable role definitions that develop over time. A "code reviewer" is not a role instantiated on demand; it is a specific agent that has reviewed hundreds of pull requests, whose \texttt{SOUL.md} has been shaped by those experiences, and whose memory index contains context that a freshly instantiated reviewer cannot access. The cost is operational: persistent agents require directory management, memory maintenance, and care about identity drift over time.

\subsubsection{SemaClaw's Current Position}

SemaClaw is committed to the persistent persona model. The design rationale is stated explicitly in Section 3.5.1: a worker agent that has handled dozens of tasks of a given type carries that history in its retrieval index, shaping the quality of its responses in ways a freshly initialized agent cannot replicate. The \texttt{SOUL.md} document is not merely a system prompt—it is a living identity anchor that accumulates and can be deliberately shaped by the operator over the agent's lifetime.

This commitment is expressed at the architectural level: the routing binding step in \texttt{resolveAgent()} is a string match against registered names, not a vector lookup against a prompt embedding pool. There are no anonymous agents in SemaClaw; every participant in a team is a registered identity with a persistent file system presence.

The DAG Teams method (Section 3.5.2) preserves the flexibility of dynamic orchestration within this constraint. The orchestrator's LLM reasons dynamically over the task and produces a dependency graph at declaration time—so the planning step retains the adaptability of a purely dynamic system—but the workers it dispatches to are registered identities, not virtual constructs. The hybrid is deliberate: dynamic planning over a static roster.

\subsubsection{What Remains Open}

The persistent persona model raises questions that the current implementation does not fully resolve.

\textbf{Identity drift over time.} An agent whose \texttt{SOUL.md} is continuously modified—by the user, by other agents, or by automated processes—may evolve in ways that invalidate the assumptions under which it was originally assigned to a role. A code reviewer whose persona document has drifted toward a general assistant is still named "code-reviewer" in the roster; the string match succeeds, but the semantic alignment the orchestrator assumed when routing to that name may no longer hold. We have no current mechanism for detecting or signaling this drift.

\textbf{Roster rigidity under novel task types.} The orchestrator routes based on what it can infer from the names and SOUL.md summaries exposed by \texttt{list\_agents}. For task types that do not map cleanly onto the existing roster, the orchestrator must either force a routing decision onto an imperfect match or declare that no suitable agent exists—there is no mechanism for composing a suitable agent on demand from the available identities. Whether this is a genuine limitation or a beneficial constraint that encourages deliberate team design is an open question.

\textbf{The case for hybrid approaches.} It is possible to imagine architectures that combine both models: a registered core of persistent specialists, augmented at task time by ephemerally instantiated generalists or role-specialized instances created from a shared base persona. The routing layer would need to distinguish between registered identities and task-time instantiations, and the memory system would need a policy for what, if anything, ephemeral agents are permitted to write. We have not explored this direction in SemaClaw, but it strikes us as a productive design space.

The deeper question is whether the choice between virtual and persistent agents is an architectural commitment or a configuration decision. SemaClaw has made it an architectural commitment; we are open to learning whether that was the right level at which to bind it.

\subsection{Harness Engineering and Model Capability — Substitution or Complement?}

\subsubsection{The Tension}

The dominant mental model for improving an AI agent system is to upgrade the underlying model. Better reasoning, larger context window, lower hallucination rates—these are properties of the model, and the conventional path to better system behavior runs through the model provider's next release. Harness engineering does not challenge this view so much as it complicates it: the question is not whether better models are better, but \textit{how much model capability is actually required to achieve a given level of system behavior}, and whether that requirement can be systematically reduced through harness design.

This is more than a cost optimization question. If harness engineering can reliably substitute for model capability on a significant class of tasks, it changes the economics of agentic deployment, the incentive structure for model development, and the competitive dynamics between large frontier models and smaller, cheaper alternatives.

\subsubsection{What the Evidence Suggests}

The case that harness engineering partially substitutes for model capability already has empirical support: the LangChain Terminal Bench 2.0 result cited in Section~1---a 13.7 percentage point completion gain from harness changes alone~\cite{LangChain2026}---is a controlled demonstration that harness investment yields capability gains independent of model investment.

SemaClaw's own architecture contains several mechanisms that operate on this principle, each reducing a different dimension of the demand placed on the base model.

\textbf{Retrieval substitutes for parametric knowledge.} A model asked to answer a domain-specific question from its weights alone must rely on what it encountered during pretraining. The same model, equipped with a well-constructed retrieval layer—FTS5 indexed memory, topic-organized Wiki entries, persona-partitioned SOUL.md context—can answer from retrieved evidence rather than parametric recall. The cognitive demand on the model shifts from \textit{knowing} to \textit{reasoning over what is provided}; the latter is a simpler task that smaller models can perform with higher reliability.

\textbf{Skill injection narrows the model's task scope.} A generalist model asked to perform a specialized task must self-calibrate its behavior from the task description alone. A model with the relevant skill injected into context—a domain-specific instruction set, example patterns, and relevant constraints loaded on demand—is asked to perform a much more constrained inference. The harness has pre-configured the cognitive context; the model executes within it rather than reasoning about it. This is the practical value of progressive skill loading: it does not extend what the model \textit{can} do so much as it concentrates the model's capability on a narrower, better-defined task where smaller models are already competent.

\textbf{Task decomposition distributes reasoning load.} The DAG Teams method (Section 3.5) breaks a complex multi-step task into a set of simpler subtasks, each assigned to an individual agent. No single model call is asked to hold the full complexity of the goal in context and reason about all of it simultaneously. A decomposed task whose hardest subtask requires only moderate reasoning depth can be served by a lightweight model throughout, where the equivalent monolithic task might have required a frontier model for the full context. The harness distributes the reasoning budget; the model only ever sees a tractable slice.

\textbf{Execution mode routing eliminates model involvement entirely for deterministic work.} The four-mode scheduled task system (Section 3.6) represents the limit case of this principle: pure script and pure notification modes consume zero tokens. For the significant fraction of scheduled work that is deterministic—data collection, file processing, status checks, recurring reminders—the harness routes the task entirely around the model. Token consumption is proportional to reasoning work, not total system activity.

\subsubsection{Implications We Are Still Working Through}

If the above observations hold consistently, they carry implications that extend beyond SemaClaw's specific design.

\textbf{A re-framing of the model selection decision.} Frontier models are optimally suited for tasks requiring genuine novel synthesis, open-ended creative reasoning, or deep multi-hop inference with no retrieval support. For the broader class of structured, context-supported, or decomposable tasks, a capable mid-tier model operating within a well-constructed harness may achieve comparable outcomes at an order-of-magnitude lower cost. The practical implication is not that frontier models are unnecessary, but that intelligent task routing—assigning model tier to task complexity rather than applying the most capable model uniformly—is a design decision with significant economic consequence. This is an argument for heterogeneous model pools managed by the harness, not for any single model.

\textbf{A possible shift in the model capability frontier that matters.} If harnesses routinely supply the knowledge (via retrieval), the structure (via decomposition), and the domain context (via skills), the residual demand on the model is primarily \textit{instruction-following reliability} and \textit{short-range reasoning coherence} rather than broad general knowledge or long-horizon planning. This suggests that, for harnessed applications, model improvements in instruction-following and structured output reliability may yield more practical return than equivalent improvements in raw benchmark performance. Whether model developers will respond to this signal—shifting optimization targets toward harness-cooperative properties—is an open question with significant implications for how the model development ecosystem evolves.

\textbf{The boundary is not fixed.} There is a category of tasks where harness substitution reaches a hard limit: tasks that require genuine novel reasoning over information that cannot be retrieved, decomposed, or pre-specified. Creative synthesis, cross-domain inference, and open-ended problem formulation cannot be fully harnessed away. The boundary between what the harness can and cannot substitute for is not static—it shifts as harness techniques improve and as the tasks users bring to agent systems evolve. Tracking this boundary empirically, rather than assuming it from first principles, seems to us a productive research direction.

\subsubsection{SemaClaw's Current Bet}

SemaClaw's design assumes that for the majority of personal productivity use cases—structured research, recurring workflows, domain-specific assistance, knowledge accumulation—a well-constructed harness can bring a capable mid-tier model to parity with a frontier model operating without harness support. We have not yet produced rigorous empirical validation of this claim in our own deployment; the LangChain data cited above is the closest external evidence we are aware of. We treat this as a working hypothesis that deserves systematic testing, and we invite the community to contribute benchmark results, failure cases, and counter-examples that would help locate where this hypothesis breaks down.

\subsection{Memory as Personal Capital — Privacy, Ownership, and the Knowledge Economy}

\subsubsection{The Tension}

The more sophisticated a memory system becomes, the more it knows about the person it serves. This is not a side effect—it is the point. A memory architecture that retains conversational history, distills recurring preferences, tracks decisions and their outcomes, and builds a topic-organized knowledge base across months of interaction is, by design, constructing a detailed and increasingly accurate model of the user: their working style, their domain expertise, their values, their relationships, their goals. The precision of this user profile is what makes the agent genuinely useful. It is also what makes the question of where that profile lives—and who controls it—a question that cannot be deferred.

The tension is fundamental: \textit{richer memory means more capable agents}, and \textit{richer memory means more exposure}. A system that knows you well enough to be useful knows you well enough to be dangerous in the wrong hands. The two properties cannot be separated at the technical level; they must be addressed at the architectural and policy level.

\subsubsection{Local Deployment as a Privacy Primitive}

SemaClaw's deployment model takes a structural position on this question before any policy is written. Memory files—\texttt{MEMORY.md}, daily conversation logs, Wiki entries, the soul directory—are stored locally, on infrastructure the user controls. There is no mandatory cloud sync, no telemetry pipeline that extracts conversational content, no model provider that receives the accumulated profile as training signal. The agent that knows you runs on your infrastructure, and the knowledge it accumulates stays there.

This is not a complete answer to the privacy question—local storage is not the same as secure storage, and a compromised local environment exposes the same data that a compromised cloud service would. But it is a meaningful architectural choice: it places the user's accumulated knowledge profile outside the attack surface of centralized service breaches, removes the incentive for the platform operator to monetize user data, and establishes a clear default that the data belongs to the user rather than to the system that generated it.

The analogy to on-premise software is instructive: the shift from SaaS to local deployment does not make data inherently safe, but it changes who bears the responsibility for protecting it and who has the power to access it. For personal AI agents that accumulate sensitive, high-fidelity user profiles, this is not a trivial distinction.

\subsubsection{Memory as Intellectual Property}

A mature memory system contains more than behavioral history. Wiki entries represent synthesized knowledge—research conducted, decisions reasoned through, domain frameworks built over time through deliberate effort. Conversation logs capture the intellectual work of formulating problems and evaluating solutions. The agent's accumulated context, in aggregate, is a record of the user's cognitive labor: the experience, expertise, and insight they have developed with the agent's assistance.

This raises a question that the current discourse around AI and privacy tends to underweight: not just \textit{how to protect} this accumulated knowledge, but \textit{whose it is} in a meaningful property sense. The knowledge in a user's memory system was produced through the interaction of the user's intentions and judgment with the agent's language capabilities. The user directed the inquiry, evaluated the outputs, corrected errors, and shaped the Wiki entries into useful form. The agent generated text. The resulting knowledge base—and the user profile implicit in it—seems to us more plausibly a form of user intellectual property than a byproduct of model inference. But no established legal or technical framework treats it as such, and the question of what rights users have over the knowledge they co-produce with AI systems is largely unresolved.

\subsubsection{The Knowledge Hub Horizon}

The skill ecosystem that has developed around OpenClaw—with ClawHub as the distribution layer—suggests a model for capability sharing that the open-source community has already validated: users build skills, share them through a common registry, and others install and benefit from them. The analogy for knowledge is natural: if skills encode \textit{how to do things}, knowledge encodes \textit{what has been learned}. A user who has spent months building a domain-specific knowledge base in SemaClaw's Wiki layer has produced something that other users in the same domain might find valuable.

A \textbf{knowledge hub} in this spirit would allow users to selectively publish portions of their accumulated knowledge base—topic trees, Wiki entries, curated memory fragments—under whatever sharing terms they choose. The spectrum of possibilities ranges from fully open (knowledge as public good, analogous to open-source code) to selectively shared (knowledge exchanged within a trusted community) to commercial (knowledge as a monetizable asset, with access gated by micropayment or subscription). The precedent for this model already exists in adjacent domains: technical documentation, research databases, educational content, and expert consulting all involve the sale or licensing of accumulated domain knowledge. The difference is that the knowledge base a personal AI agent produces is continuously updated, highly personalized, and structurally ready for machine consumption.

The compounding property is worth noting explicitly. Unlike skills, which tend to be relatively static once authored, a living knowledge base accretes value over time as new interactions refine existing entries, fill gaps, and connect previously isolated concepts. A user's knowledge base after two years of active use is meaningfully more valuable than after six months—not just to the user, but potentially to others who want to benefit from that accumulated expertise. This creates an incentive structure for sustained knowledge curation that has no direct analogue in the current open-source ecosystem.

\subsubsection{What Remains Open}

The knowledge hub vision raises questions that technical design alone cannot resolve.

\textbf{Provenance and attribution.} When a user publishes a Wiki entry, how much of its content is the user's intellectual contribution and how much is the model's generated text? What attribution is owed to the model provider whose inference produced the raw material? These questions do not have established answers, and the emergence of knowledge marketplaces would force them into the open.

\textbf{Containment and leakage.} A user who selectively publishes portions of their knowledge base must trust that the published portions do not implicitly reveal what they intended to keep private. Topic-organized Wiki entries can be revealing in their structure even when their content is scrubbed: the \textit{existence} of a detailed knowledge tree in a particular domain says something about the user's interests and activities. Granular access control at the entry level may be insufficient; the question of what structural metadata is safe to expose alongside published content is non-trivial.

\textbf{Trust and quality in a knowledge marketplace.} Skills in ClawHub are executable—users can test whether a skill works before relying on it. Knowledge entries are not directly testable in the same way; their value depends on accuracy, currency, and relevance, which are harder to verify at installation time. A knowledge hub would need reputation mechanisms, versioning, and possibly domain-specific quality signals that the current skill ecosystem does not require.

These questions are not reasons to abandon the knowledge hub direction—they are the design problems that would need to be solved to realize it responsibly. We surface them here because we think they represent some of the most interesting and consequential open problems in the personal AI agent space, and because the community building in this direction will need answers that do not yet exist.

\subsection{The Next Forms of Harnessed Plugins}

\subsubsection{Beyond Tool Use}

The four-layer plugin taxonomy introduced in Section 2.3—MCP tools, subagents, skills, hooks—was designed to describe the existing landscape of agent capability extension and map each layer onto the engineering phase it primarily operates within. MCP tools extend the action space at the ReAct tool-use layer. Subagents delegate reasoning through prompt-defined interfaces. Skills inject domain context on demand. Hooks insert control logic at lifecycle events. The taxonomy is descriptive and useful, but it is organized around \textit{where} extensions operate rather than \textit{what they produce}.

As harness engineering matures, a new design question comes into focus: what happens when a plugin is designed not just to extend what an agent can do in a single session, but to \textit{change what the agent is} over time? The existing taxonomy has no clean category for this. A plugin that modifies the agent's memory, evolves its persona, or produces artifacts that future agent sessions will rely on is doing something qualitatively different from a tool that fetches data or a skill that loads a prompt template. It is operating at the level of the harness infrastructure itself—reading from and writing to the persistent state that the harness manages.

The wiki-based personal knowledge infrastructure described in Section~3.7 is SemaClaw's first attempt to build in this direction. It is worth being precise about why this knowledge layer represents a new category rather than just another tool.

\subsubsection{Wiki-Based Personal Knowledge Infrastructure as a Design Prototype}

A conventional MCP tool is stateless with respect to the agent's long-term architecture: it accepts inputs, produces outputs, and leaves the agent's memory, persona, and context infrastructure exactly as it found them. The current implementation of this knowledge layer (Section~3.7) does not work this way---its outputs are not merely returned to the user but written into the persistent corpus that future agent sessions will retrieve from. Its effect is not just a response; it is a mutation of the agent's own long-term knowledge infrastructure.

This makes the current implementation a \textit{stateful harness plugin}: its effects persist beyond the session in which it is invoked, shaping the agent's future capabilities in ways that accumulate over time. The design implications are significant. A stateless tool can be invoked freely with no concern for side effects on the agent's architecture; the worst case is a bad output that the user ignores. A stateful harness plugin changes persistent state, and a poorly designed one can introduce noise, contradictions, or structural drift into this knowledge layer that future sessions will operate on. This demands a different design standard: not just ``does this produce a useful output'' but ``does this leave the agent's persistent architecture better than it found it.''

\subsubsection{Forms We Have Not Yet Built}

Among the stateful harness plugin forms we have identified, this knowledge layer is the most basic instantiation of the concept. Several more sophisticated forms suggest themselves, each raising distinct design challenges.

\textbf{Persona evolution plugins.} A \texttt{SOUL.md} document is seeded at agent creation and shaped by the operator over time—but the shaping is manual. A persona evolution plugin would allow an agent to propose revisions to its own SOUL.md based on patterns it has observed in its interactions: recurring task types it has handled, domains it has developed depth in, working styles that have proven effective. The agent is not merely using its persona—it is participating in its own identity development. The design challenge is containment: a plugin with write access to SOUL.md can destabilize the identity anchor that the orchestrator's routing decisions depend on. Gating proposed revisions through a human-approval step is the obvious safeguard, but the interaction design for that approval flow is non-trivial.

\textbf{Cross-agent knowledge plugins.} In a multi-agent team, each agent's memory is isolated under its own folder. A cross-agent knowledge plugin would allow an agent to write a distilled finding into a shared team knowledge space that other agents can retrieve from. This is a form of structured inter-agent communication that goes beyond the task-level message passing that \texttt{dispatch\_task} currently supports: instead of passing a result from one task to the next, the agent is contributing to a persistent shared context that the entire team reads from over time. The memory architecture, currently organized around individual agent namespaces, would need a shared-namespace layer to support this, along with policies for write access, conflict resolution, and namespace hygiene.

\textbf{Evaluation and self-correction plugins.} A plugin that, after a task completes, compares the agent's output against a stored quality criterion and writes the verdict—along with the failure mode, if any—into a structured evaluation log would enable systematic improvement tracking over time. This is not a new idea in the ML literature, but implementing it as a harness plugin rather than as an external evaluation pipeline changes its availability: it runs inside the agent's own execution context, with access to the full reasoning trace, and its outputs live in the agent's memory alongside the work it is evaluating. The challenge is avoiding evaluation drift—a system that writes its own evaluations can gradually normalize mediocre performance if the quality criteria are not anchored to an external standard.

These examples are sketches, not a roadmap. The stateful harness plugin space is open-ended by design, and its eventual shape will be drawn by what the community builds rather than by what we anticipate here.

\subsubsection{The Design Principle at Stake}

What these forms share is a common property: they are plugins that \textit{operate on the harness} rather than plugins that the harness merely \textit{executes}. The distinction matters because it requires a different security model. A plugin with write access to persistent state—memory, persona, shared knowledge—has leverage over all future agent behavior that depends on that state. This is a significantly higher level of trust than a plugin that fetches a web page or runs a calculation.

The practical implication is that plugin permission tiers need to distinguish not just between safe and high-risk \textit{actions} (the current PermissionBridge model) but between plugins that operate within session scope and plugins that operate on persistent infrastructure. SemaClaw has not yet built this distinction into its permission architecture. The current implementation is deployed with the implicit assumption that the user trusts it to write to their knowledge base---a reasonable assumption for a first-party capability, but one that becomes fragile as a community plugin ecosystem grows and third-party plugins acquire write access to the same infrastructure.

This is the frontier that the knowledge layer opens rather than closes: not just ``what should agents be able to do'' but ``what should plugins be able to do to the agent itself.''

\subsection{Beyond the Individual Agent — Community Forms in the Harness Era}

\subsubsection{A Shift in the Unit of Analysis}

The preceding questions have all taken the individual agent—or a team of agents under a single operator's control—as the unit of analysis. The system prompt, the memory, the persona, the skill set: these are all configured by and for one user. This is the natural starting point for personal AI agent design, and it is where most of the current engineering work is concentrated.

But the open-source agent ecosystem is already producing community forms that do not fit this model. ClawHub is the clearest example: a registry where individual users publish skills they have authored, and other users install and build on them. The unit of analysis here is not the individual agent—it is the community of agents whose capabilities are shaped by a shared, evolving skill commons. Skills authored by one user's agent workflow become capabilities available to thousands of others. The individual and the collective are linked through a shared infrastructure layer.

This pattern, already established in the skill dimension, raises a larger question: as agent systems mature and proliferate, what other community forms will emerge—and what engineering infrastructure do they require?

\subsubsection{Agent-to-Agent Interaction}

The most technically immediate extension is \textbf{agent-to-agent (A2A) communication} across operator boundaries. SemaClaw's DAG Teams model enables one agent to delegate to another within a single deployment. The next step—an agent delegating to, querying, or collaborating with an agent operated by a different user—requires solving problems that intra-system orchestration does not face.

\textit{Identity and authentication.} Within a single deployment, agent identity is a string match against a local registry. Across deployments, an agent claiming to be "research-assistant operated by user X" needs a credential that the receiving system can verify. The model here is closer to service-to-service authentication in distributed systems than to the current local-roster lookup.

\textit{Trust propagation across operator boundaries.} Within a team, the orchestrator's authority is inherited by the workers it dispatches to—SemaClaw's current model implicitly trusts the orchestrator's task assignments. When the orchestrator is another user's agent operating remotely, this implicit trust does not carry. The receiving system must decide how much authority to grant an external agent's instructions, under what conditions, and with what recourse when instructions turn out to be harmful. The attack surface around trust propagation becomes significantly more complex when the orchestrator-worker relationship spans different operators and potentially different security boundaries.

\textit{Capability discovery.} For one agent to usefully delegate to another, it needs to know what the other can do. This is a capability advertisement problem that the current skill registry partially addresses at the human-to-agent layer—a user can browse ClawHub to find skills that extend their agent's capabilities—but does not address at the agent-to-agent layer. An agent that can autonomously discover, evaluate, and invoke the capabilities of other agents requires a structured capability description format and a discovery protocol that does not yet exist in the open-source ecosystem.

\subsubsection{Human-Agent Community Forms}

The more subtle shift is in how \textit{humans} organize around agents. ClawHub established the skill-sharing community; platforms like Moltbook~\cite{moltbook2026} are exploring a different form altogether: a social network in which agents are the primary participants---posting, commenting, upvoting, and building communities---while humans participate as owners who claim, observe, and manage their agents through a paired dashboard. The unit of community here is not a shared configuration but a shared social space: agents interact across operator boundaries, and humans follow the activity of agents they trust or find interesting.

Several community forms seem plausible as this direction develops:

\textbf{Shared agent templates.} A user who has configured a highly effective research agent—SOUL.md, memory organization, skill loadout, scheduled task patterns—could publish the configuration as a template that others can instantiate and personalize. The value is not the agent's accumulated memory (which remains private and user-specific) but the \textit{structural knowledge} of how to set up an effective agent for a given use case. This is knowledge about agent design rather than knowledge produced by the agent.

\textbf{Domain-specialized agent communities.} Users who work in the same domain—legal research, software architecture, scientific literature review—face similar agent configuration challenges and accumulate similar types of knowledge. Communities organized around shared domain contexts could develop shared vocabulary for agent persona design, shared skill ecosystems tuned to domain-specific tools and sources, and shared evaluation criteria for what "good" looks like for an agent in that domain. ClawHub's skill taxonomy already begins to organize along these lines; a more deliberate community design could accelerate it.

\textbf{Collaborative knowledge commons.} Seciton~4.3 introduced the knowledge hub concept as a mechanism for individual users to share or monetize accumulated knowledge. A further step is collective knowledge production: communities of agents whose users have opted into a shared knowledge pool, where individual agents contribute findings that the entire community benefits from. The model is closer to a cooperative research network than to a marketplace—contributions and benefits are reciprocal rather than transactional. The governance challenges of such a commons—what counts as a valid contribution, how conflicts between contributed entries are resolved, how quality is maintained over time—are significant and largely uncharted.

\subsubsection{The Infrastructure Gap}

What is notable about these community forms is that none of them are fully supported by the current open-source agent infrastructure. ClawHub provides skill distribution; it does not provide agent-to-agent authentication, capability discovery, cross-operator trust management, or the shared knowledge infrastructure that a knowledge commons would require. Moltbook and similar platforms are early experiments in A2A social infrastructure, not mature platforms. The community forms that seem most valuable are the ones that are furthest from being buildable today.

This is, in our view, the most consequential long-horizon engineering question in the personal AI agent space: not how to make an individual agent more capable, but how to build the infrastructure that allows communities of agents—and communities of humans organized around agents—to produce collective intelligence that no individual agent could produce alone. The harness engineering work described in this paper is a precondition for that infrastructure, not the infrastructure itself. Taken across all five questions in this section, SemaClaw's contribution is to demonstrate what is possible at the individual-agent level; the community forms that become possible on top of it are a problem for the ecosystem to solve together.

\section{Conclusion}
This paper has made a single sustained argument: that the reliability, safety, and long-term usefulness of a personal AI agent system are primarily engineering properties—determined by the structure of the harness around the model, not by the model alone—and that building in this direction requires a coherent, layered architecture where each design decision reinforces the others.

\subsection{The System as a Whole}

\begin{figure}[t]
    \centering
    \includegraphics[width=\columnwidth]{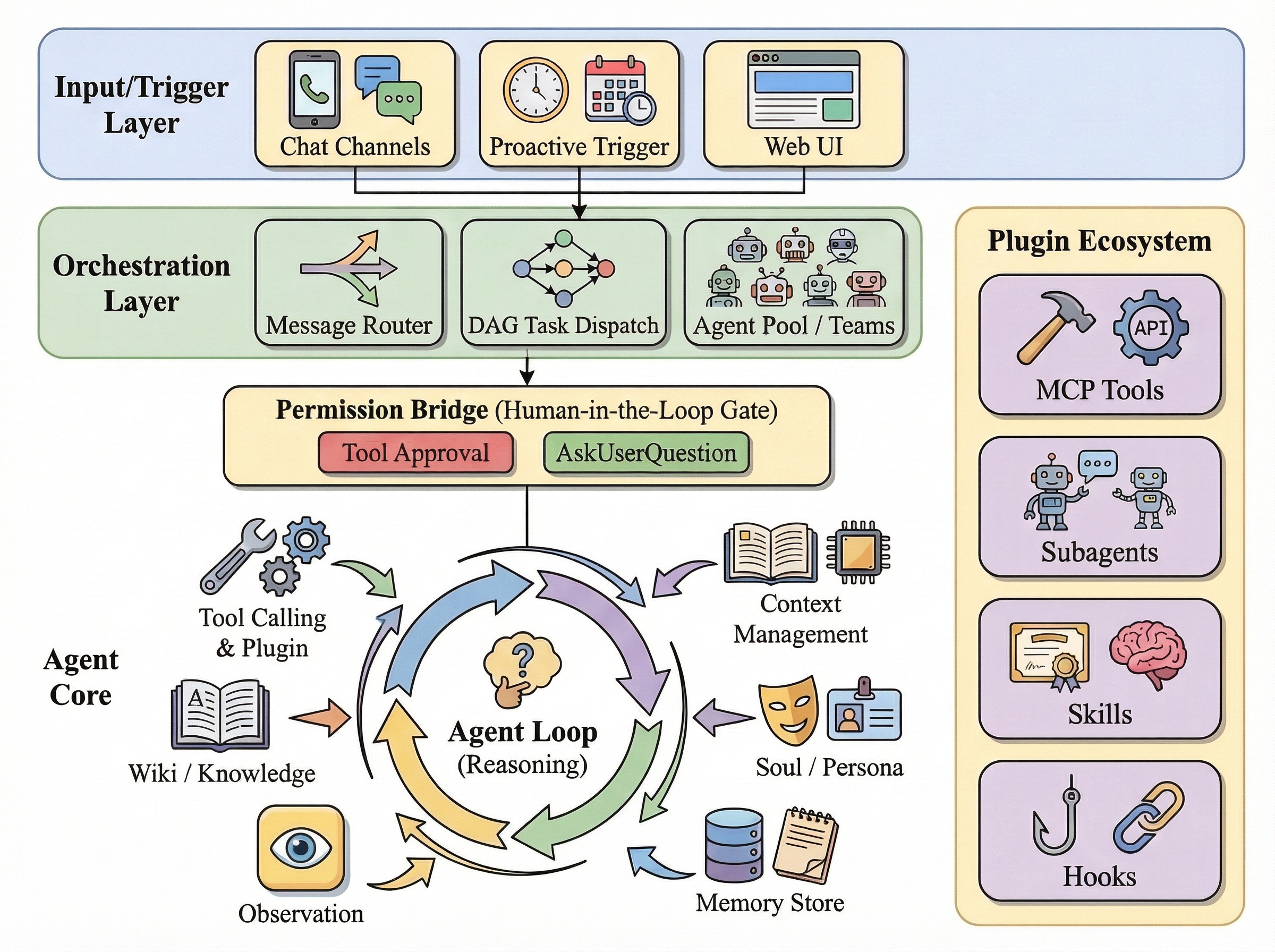}
    \caption{SemaClaw's full architecture stack: seven contributions forming a coherent harness from runtime foundation to application surface.}
    \label{fig:product_arch}
\end{figure}

The seven architectural contributions described in Section 3 are not independent features. They form a stack in which each layer depends on the integrity of the layers beneath it. Figure~\ref{fig:product_arch} maps the resulting structure.

At the foundation, the separation of \texttt{sema-code-core} from \texttt{semaclaw} establishes a clean runtime boundary: the execution loop, tool orchestration, and multi-tenant isolation are encapsulated in a reusable library, while the application layer adds channel integration, message routing, and persona management on top. This boundary makes each layer independently evolvable and the runtime reusable beyond any single application context.

Above the runtime, the three-tier context architecture---working memory with compaction, external memory with hybrid retrieval, and structured context injection spanning soul, workspace, and rules---gives each agent a stable identity, cross-session recall, and task-specific context as independently governed layers. The PermissionBridge sits as a horizontal gate across all agent execution: a human-in-the-loop mechanism that applies the principle of least privilege at the tool boundary without interrupting session continuity. The plugin ecosystem—MCP tools, subagents, skills, and hooks—defines the extension surface through which any of these capabilities can be augmented or composed without modifying the runtime itself. Agent Teams add a coordination layer above the individual agent, enabling declarative DAG-driven multi-agent execution grounded in persistent personas rather than ephemeral role assignments. Scheduled tasks extend the system's temporal reach, routing each job to the appropriate execution primitive based on its complexity class. The wiki-based personal knowledge infrastructure closes the loop between task execution and durable knowledge: intelligence produced through interaction is captured as a user-owned Markdown corpus the user can browse, edit, and reorganize, and from which future agent sessions retrieve as a curated knowledge layer.

Together, these layers instantiate the harness engineering model for the specific domain of general-purpose personal AI assistants. The result is a system that is structurally safe by default—agent authority is bounded, human oversight is native to the execution path, and each extension point is governed—while remaining open and composable at every level. The architecture accommodates a range of deployment contexts, from a single user running a personal assistant for daily knowledge work to a small team operating a standing roster of specialized agents on shared projects. In both cases, the same layered harness governs execution, and the same mechanisms accumulate value over time.

The architecture described above also addresses the second arc introduced in Section~1: the evolution of human--agent interaction from message-level exchange toward a persistent collaborative relationship. This evolution places two demands on the system that purely capability-focused frameworks do not meet. The first is \textit{trustworthy delegation}: as the interaction unit shifts from a message to a goal, users extend real authority to agents acting on their behalf across files, APIs, and external services. PermissionBridge is SemaClaw's answer to this demand---not as a safety add-on, but as a native runtime primitive that makes goal-level delegation structurally safe, allowing users to expand the agent's trusted action space incrementally rather than all at once. The second is \textit{relational continuity}: a persistent collaborative relationship requires an agent that remembers, accumulates, and organizes---one whose usefulness compounds over time rather than resetting at each session boundary. The three-tier context architecture provides the memory substrate for this continuity; the wiki-based personal knowledge infrastructure is its active expression, transforming each completed interaction into a contribution to a growing body of shared knowledge. Together, these two mechanisms are what separates a capable task executor from a genuine long-term collaborator.

\subsection{Contributions}

SemaClaw's concrete contributions to the open-source agent ecosystem are as follows.

\textbf{A harness architecture for personal AI.} The full SemaClaw framework demonstrates how the foundational capabilities of \texttt{sema-code-core}—context management, tool use, compaction—can be composed with channel integration, persistent memory, permission gating, and multi-agent coordination into a coherent application harness. The architecture is documented, open-source, and designed to be forked, extended, and adapted.

\textbf{Operational primitives that are missing from the current ecosystem.} Several mechanisms described in this paper—the PermissionBridge human-in-the-loop gate, the persistent-persona agent team model with DAG-driven dispatch, the four-mode scheduled task system, and the stateful harness plugin design represented by the wiki-based personal knowledge infrastructure—address engineering problems for which no widely adopted open-source solution currently exists. We offer these as starting points for community convergence rather than finished answers.

\textbf{A vocabulary for harness engineering in this domain.} The framework presented in Section 2—the cognitive architecture typology, the four-layer plugin taxonomy, the orchestration strategy comparison, the persona context model—attempts to give the community a shared vocabulary for discussing design decisions that are currently made by convention or intuition. Shared vocabulary is a precondition for shared evaluation criteria, and shared evaluation criteria are what eventually separate engineering from craft.

\subsection{Limitations}

SemaClaw's current implementation has limitations that its architecture does not resolve. First, the interaction surface currently centers on CLI and Web UI; integration with broader communication channels—messaging platforms, email, voice interfaces—remains incomplete, limiting the agent's reach in non-development scenarios. Second, the broader AI-agent plugin ecosystem is evolving rapidly; SemaClaw has not yet fully aligned with Claude Code's upstream extension model—notably, user-defined hooks are not yet integrated, and divergences in permission semantics and context conventions may require ongoing reconciliation as upstream capabilities expand. Third, the Wiki knowledge base operates as a standalone retrieval surface; deeper integration with the agent's memory search pipeline—enabling cross-source retrieval that spans both structured wiki content and session-derived memory—is a natural next step that the current architecture supports but does not yet implement.

These scope boundaries are characteristic of a first-generation system and define a clear roadmap for subsequent development.

\subsection{Looking Forward}

The open questions raised in Section 4—orchestration architecture, harness-model complementarity, memory ownership, stateful plugin design, and community infrastructure—mark the boundary of what current systems, including SemaClaw, have solved. They are not edge cases; they are the next tier of problems that the field will need to work through as personal AI agents move from early adoption to broad deployment.

SemaClaw's contribution is to demonstrate what a layered, harnessed personal AI system looks like in practice, to make the implementation available as a reference and a starting point, and to surface the engineering decisions that required explicit choices so that the community can evaluate and improve on them. The ambition is not to define the final architecture of personal AI agents—it is to advance the shared engineering foundation on which better architectures can be built.

\clearpage
\newpage
\bibliographystyle{assets/plainnat}
\bibliography{paper}


\end{document}